\documentclass{article}

\PassOptionsToPackage{numbers, compress}{natbib}
\usepackage[final]{neurips_2023}

\usepackage[utf8]{inputenc} 
\usepackage[T1]{fontenc}    
\usepackage[pagebackref,breaklinks,colorlinks]{hyperref}
\hypersetup{ colorlinks=true, linkcolor=blue, filecolor=magenta, urlcolor=cyan, }
\usepackage{url}            
\usepackage{booktabs}       
\usepackage{amsfonts}       
\usepackage{nicefrac}       
\usepackage{microtype}      
\usepackage[dvipsnames,table]{xcolor} 
\usepackage{xspace}
\usepackage{graphicx}
\usepackage{amssymb}
\usepackage{pifont}
\usepackage{duckuments}
\usepackage{enumitem}
\usepackage{amsmath}
\usepackage{multirow}
\usepackage{float}
\usepackage{afterpage}
\usepackage{adjustbox}
\usepackage{algorithm}
\usepackage{algpseudocode}
\usepackage{caption}
\usepackage{subcaption}
\usepackage{hyperref}
\usepackage{cleveref}
\usepackage{listings}
\usepackage[symbol]{footmisc}
\renewcommand{\thefootnote}{\fnsymbol{footnote}}

\lstdefinestyle{mystyle}{
    basicstyle=\ttfamily\small,
    breaklines=true,
    frame=single,
    backgroundcolor=\color{white},
    columns=fullflexible,
    keepspaces=true,
    showstringspaces=false,
    tabsize=4,
    breakatwhitespace=true,
    breakindent=0pt,
}

\usepackage{placeins} 
\usepackage{titletoc} 

\newcommand{\ie}{\textit{i.e.}}
\newcommand{\eg}{\textit{e.g.}}

\definecolor{baselinecolor}{gray}{.9}

\definecolor{deemph}{gray}{0.7}

\definecolor{ablationblue}{RGB}{20,124,234}
\definecolor{ablationgreen}{RGB}{20,132,109}

\title{ViPer: Visual Personalization of Generative Models via Individual Preference Learning}

\author{
\textbf{Sogand Salehi} \quad \textbf{Mahdi Shafiei} \quad {Teresa Yeo}  \vspace{0.3em} \\
\textbf{Roman Bachmann} \quad \textbf{Amir Zamir} \vspace{0.3em} \\ \vspace{1em}
Swiss Federal Institute of Technology Lausanne (EPFL) \\ \url{https://viper.epfl.ch} \\
}
\begin{document}

\renewcommand{\thefootnote}{\fnsymbol{footnote}}
\setcounter{footnote}{1}

\maketitle
\begin{figure}[H]
  \centering
  \includegraphics[width=\linewidth]{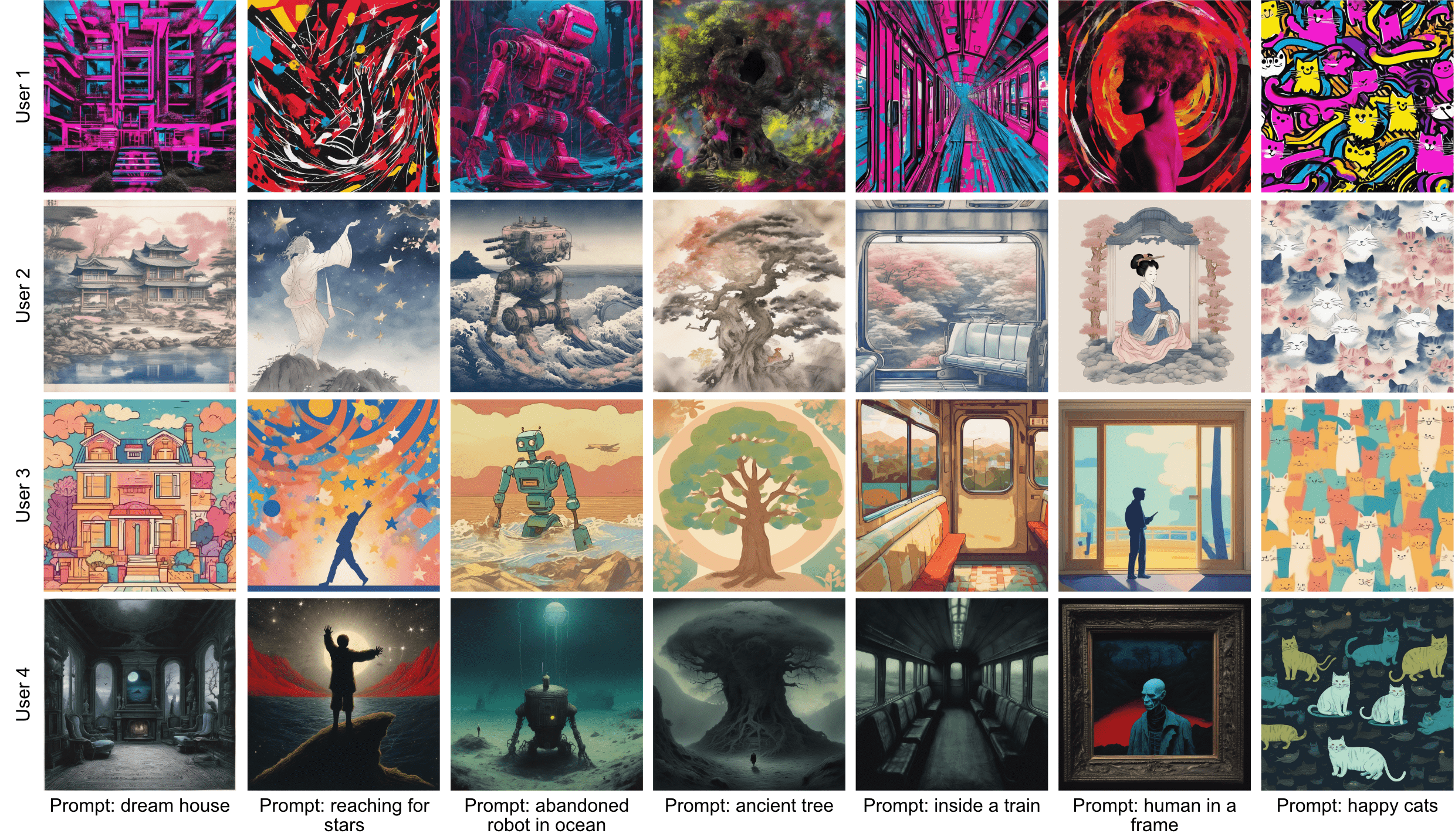}
  \caption{We introduce ViPer, a method that \textit{personalizes} the output of generative models to align with different users' preferences \textit{for the same prompt}. This is done via a one-time capture of the user's general preferences and conditioning the generative model on them without the need for engineered detailed prompts. Notice how the results vary for the same prompt for different users based on their visual preferences.}
  \label{fig:pull_figure}
\vspace{-1em}
\end{figure}

\begin{abstract}
Different users find different images generated for \textit{the same prompt} desirable. This gives rise to \textit{personalized} image generation which involves creating images aligned with an individual’s visual preference. 
Current generative models are, however, unpersonalized, as they are tuned to produce outputs that appeal to a broad audience. Using them to generate images aligned with individual users relies on iterative manual prompt engineering by the user which is inefficient and undesirable.

We propose to personalize the image generation process by first capturing the generic preferences of the user in a one-time process by inviting them to comment on a small selection of images, explaining why they like or dislike each.
Based on these comments, we infer a user’s structured liked and disliked visual attributes, \ie, their visual preference, using a large language model. These attributes are used to guide a text-to-image model toward producing images that are tuned towards the individual user's visual preference.
Through a series of user studies and large language model guided evaluations, we demonstrate that the proposed method results in generations that are well aligned with individual users' visual preferences.

Our code and model weights are open sourced at \url{https://viper.epfl.ch}.

\end{abstract}


\section{Introduction}
\label{sec:intro}
Text-to-image generative models \cite{rombach2022highresolution, ho2020denoising} have demonstrated the ability to produce high-fidelity images and exhibit remarkable generalization abilities. This capability has sparked considerable interest regarding their potential applications which include image editing~\cite{kawar2023imagic, parmar2023zeroshot, brooks2023instructpix2pix, mokady2022nulltext}, style transfer~\cite{he2024freestyle, chung2023style, ruta2023diffnst}, and generating training data~\cite{he2022synthetic,sariyildiz2023fake,luo2024scdiffusion,yeo2024controlled}.

These applications generally require guiding the generative model towards certain desired outputs. 
In this work, we are interested in guiding the generations toward an individual's preferences in images, or \textit{personalization}.

Current methods for producing personally desirable images require the users to iteratively refine their prompts until they achieve their preferred outcomes or incorporate sparse feedback. Some forms of this feedback include binary reactions \cite{vonrütte2023fabric}, providing images that the individual likes or dislikes \cite{sohn2023styledrop, gal2022image, ruiz2023dreambooth}, or ranking a given set of images \cite{tang2024zerothorder}.

We propose to utilize the individuals' comments on a given set of images for capturing their generic visual preferences. This allows them to express \textit{why} they like or dislike certain images. We will demonstrate, through user studies, that this increased level of expressivity results in generations better aligned with individual preferences. To incorporate this feedback into a text-to-image model like Stable Diffusion~\cite{rombach2022highresolution, podell2023sdxl}, we first transform these comments into a structured representation containing visual attributes, \ie, a set of features they like or dislike in images, \eg, surrealism, rough brushstrokes, monochromatic, etc. We demonstrate that we can condition Stable Diffusion on these visual preferences without any additional fine-tuning. We call our method Visual Personalization of Generative Models via Individual Preference Learning, or ViPer for short.

Our user studies show that users prefer our method over other baselines most of the time. They also strongly favor personalized results tailored to their preferences compared to non-personalized results or results from other users.
\section{Related Work}

\textbf{Aligning image generative models with \textit{general} human preferences.} Generative models such as Stable Diffusion~\cite{rombach2022highresolution} are trained on images from the internet. Thus, their generations may reflect the artifacts of such data \eg, unusual object poses. This is not ideal if the goal is to generate aesthetically pleasing images \eg, high quality images where the object is centered and unoccluded.
To align image generative models with these general goals, as opposed to an individual's preference, these models are often tuned with reinforcement learning from human feedback (RLHF) \cite{wallace2023diffusion, clark2023directly, prabhudesai2023aligning} to maximize scoring metrics \cite{wu2023human, kirstain2023pickapic, xu2023imagereward, AestheticsLaion2022} and improve image quality. \cite{hao2023optimizing, wen2023hard} optimize input prompts instead to achieve the same. While these methods enhance the quality of results by making them aesthetically aligned with general human preferences, ViPer focuses on an individual's preferences and generates personalized images.

\textbf{Aligning image generative models with \textit{individual's} preferences.}
Personalization methods seek to adapt generative models to specific concepts, such as styles, by leveraging feedback or direct inputs from individuals. Popular examples include \cite{gal2022image} and \cite{ruiz2023dreambooth}, which aim to learn a representative token for desired concepts, and guidance methods \cite{song2021scorebased, dhariwal2021diffusion, ho2022classifierfree, li2022upainting}, which steer sampling towards images that satisfy a desired objective by adding an auxiliary term to the score function. \cite{sohn2023styledrop} learns a new style by fine-tuning a few trainable parameters from a reference image and improving the quality via iterative training with either human or automated feedback while \cite{vonrütte2023fabric} is a training-free method that incorporates iterative feedback into the generation process of text-to-image models. These methods mostly rely on users to express their preference for a set of images by simply liking or disliking them, or by providing images themselves. In contrast, ViPer allows users to express their detailed opinions about a given set of images to generate personalized results. This is a one-time process and users aren't required to provide further feedback. In~\cref{sec:userevalbaselines}, we show that the additional flexibility and expressivity of comments result in generations more aligned with users' preferences.

\begin{figure}[tb]
  \centering
  \includegraphics[width=\linewidth]{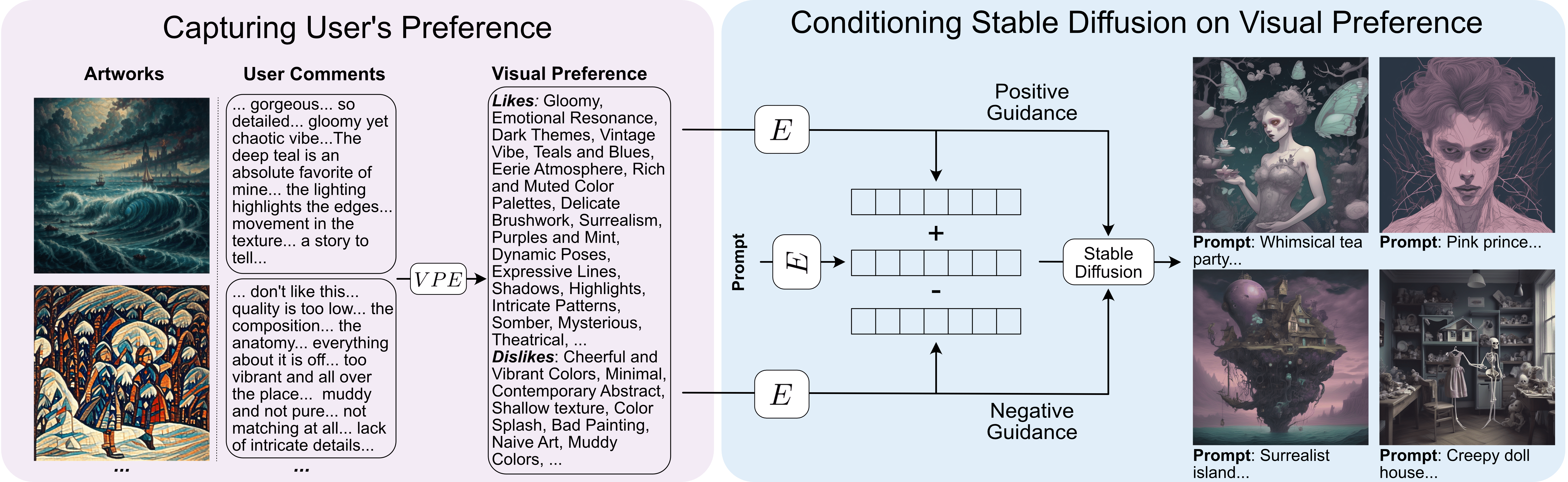}
  \caption{
  \textbf{(Left) Capturing an individual's preference from comments.} 
  We ask interested individuals to comment on a small set of images. 
  These images are generated such that they have diverse styles (see \cref{sec:commentimages} for visuals and further details). Note that these comments are not required to be of a specific structure. Thus, users can write as much or as little as they like. Obviously, more detailed and expressive comments would lead to better personalization. We show an example of an individual's comments on two images. Our method allows for such free-form comments as we make use of a language model to extract structured preferences, i.e. a visual preference extractor ($VPE$). To get the $VPE$, we fine-tune IDEFICS2-8b. Converting the user's free-form comments into structured visual preferences is a one-time process that provides us with a concise representation of the user's preferences. In this example, $VPE$ extracts the individual's preferences for the color palette, vibe, and lighting from the comments and translates them into attributes like "Rich and Muted Color Palettes", "Gloomy", and "Highlights and Shadows". \textbf{(Right) Conditioning Stable Diffusion on an individual's visual preference.} The user's visual preferences are then encoded and added to the prompt embedding (see~\cref{eq:embeddingaddtion}). This allows us to steer the generations towards certain styles. They're also used in the guidance formula (see~\cref{eq:guidance}) directly to guide Stable Diffusion's results toward the user's preferences. This step does not require any fine-tuning and can be used directly with the Stable Diffusion model. Note how the generations reflect the user's preference by generating a "Gloomy" style while avoiding styles like "Contemporary Abstract", "Vibrant Colors", and "Naive Art".}
  \label{fig:overview_figure}
\end{figure}

\section{Method}
We propose a method to align text-to-image models with an individual's visual preferences.
In~\cref{sec:method-vp}, we propose an approach for \textit{computing a signal that captures an individual's preferences}.
In~\cref{sec:method-personalize}, we demonstrate how we can \textit{use this signal to adapt a generative model} to generate personalized images. 
Finally, in~\cref{sec:method-eval}, we introduce a proxy metric to evaluate \textit{how well-aligned} the personalized images are with the individuals' preferences. 
See~\cref{fig:overview_figure} for an overview of our method.

\subsection{Capturing an individual’s visual preferences}\label{sec:method-vp}
Current methods for capturing preferences include utilizing signals such as whether an individual likes or dislikes a set of images \cite{vonrütte2023fabric}, or how they rank each image within the set \cite{tang2024zerothorder}, while some require users to provide a set of representative images \cite{gal2022image, ruiz2023dreambooth, sohn2023styledrop}. However, these \textit{signals are sparse} and may only cover some aspects of an individual's preferences. 

Furthermore, our preferences and even the images themselves can be complex, making it challenging for these signals to fully capture the intricacies of an individual's preferences. Take for example someone who likes cool colors and realistic art and dislikes warm colors and pixel style. In traditional binary choice or ranking methods, when presented with pixel art featuring cool colors, the individual must decide whether they like the image despite the pixel style, or rank how much they like it.

\textbf{Capturing an individual's preferences with free-form comments.} We propose to use \textit{comments} on a given set of images to capture an individual's preferences. To do so, we first generate around 50 images with diverse art features such as style (see the \cref{sec:commentimages} for visuals) and ask individuals to select a random subset to comment on. Note that one can also comment on their own provided set of images if they prefer. Individuals are encouraged to discuss \textit{why} they like or dislike each selected image, their perception of it, and the emotions it evokes. Importantly, individuals are not bound to any specific writing structure when commenting. They can write as much or as little as they prefer since we \textit{utilize a language model to extract structured preferences from their free-form comments.}
Ideally, these comments are descriptive, covering most of the attributes present in the images that are important to the individual. We show that in this case, comments on as few as eight images are enough to get personalized results~(see \label{sec:numcom} for examples.)

\textbf{Converting free-form comments to a structured representation of an individual's preferences.} We propose representing an individual's visual preferences through a set of attributes across 20 categories such as texture and color palette. Each category contains multiple attributes, \eg within the texture category, attributes include smoothness, roughness, etc. To obtain this, we instructed several language models \eg, GPT-3.5\footnote{openai.com}, Claude 2.1\footnote{claude.ai}, and Llama 2 to output representative art categories and their corresponding attributes. By aggregating their results, we compiled a total of approximately 500 visual attributes. See page \pageref{visat} for the complete list.

To transform the free-form comments into visual preferences, we require a dataset that maps comments and individual preferences. Since such data does not exist to our knowledge, we start by creating 5000 agents, or simulated individuals, who are characterized by their liked and disliked visual attributes. Each agent is randomly allocated an average of 50 liked and disliked attributes from the attributes list mentioned above. To ensure diversity, we generate them such that there is a minimum Jaccard distance of 85\% between any two agents' liked and disliked attributes.

Next, we generate comments that each agent would write for various images based on their visual preferences. To keep the comments relevant to visual preferences and informative, we first generate an average of 10 images for each agent, concatenating a random subset of visual attributes they like or dislike in the input prompts. Although the images are still somewhat random since the input prompts are chosen randomly, this set is directly related to each agent's visual preferences.

To generate the agents' comments on these images, we instruct GPT-4~\cite{openai2024gpt4} to create each agent's comments based on their visual preferences. Typos and sarcasm are randomly allowed to add diversity and make the results more human-like. This process results in a \textit{paired dataset of visual attributes and comments} for each agent.

To extract an individual's visual preference from their comments, we fine-tune IDEFICS2-8b \cite{laurencon2023obelics, laurençon2024matters}, an open-sourced vision-language model, on the agents' paired dataset. We found this model to work best compared to other open-source vision-language models. We demonstrate the shortcomings of our initial experiments with other models in \cref{sec:vlm}. We denote the fine-tuned IDEFICS model as the Visual Preference Extractor, or $VPE$. 
Note that the $VPE$'s output is {not limited to the pre-defined attributes}; it can extract attributes mentioned in comments, \eg, recognizing a color mentioned in the comments but not defined in our attribute set.

Instead of converting comments to visual preferences, one might consider directly asking individuals to choose from visual attributes. However, we believe that this process can be tedious due to the large number of attributes (about 500) and certain attributes may be challenging to visualize out-of-context.

\subsection{Personalizing a generative model based on an individual’s preference} \label{sec:method-personalize}
In the previous section, we described how we can capture an individual's visual preference from their comments. Here, we are interested in how we can use these visual preferences to adapt a generative model. In this work, we employ Stable Diffusion~\cite{rombach2022highresolution, podell2023sdxl}. As described previously, each individual's preferences are described by a set of preferred and disliked attributes, denoted as $VP_+$ and $VP_-$ respectively. Encoding these preferences using Stable Diffusion's text encoder, $E$, gives us $E(VP_+)$ and $E(VP_-)$. We encode the input prompt, $p$, and modify the resulting embedding, $\mathbf{p}$, with $E(VP_+)$ and $E(VP_-)$ to steer the generations towards the individual's preferences as follows:
\begin{align}
    \mathbf{p} = E(p) + \beta (E(VP_+) - E(VP_-)). \label{eq:embeddingaddtion}
\end{align}
$\beta$ can be tuned by the individual to increase or decrease the `level' of personalization according to their preferences. See~\cref{fig:pers_degree} for examples. We found in our experiments that $\beta\leq 1$ ensures that the generations will be faithful to both the input prompt and visual preferences. To guide the process toward the individual's preferences along with the prompt even further, we also modify the predicted noise by using classifier-free guidance~\cite{ho2022classifierfree, liu2023compositional}. (see~\cref{eq:guidance}). For a given timestep $t$,
\begin{align}
    \mathbf{\epsilon}_{vp}(\mathbf{x}_{t}, t) &= \mathbf{\epsilon}_{\theta}(\mathbf{x}_{t}, t, E(VP_+)) - \mathbf{\epsilon}_{\theta}(\mathbf{x}_{t}, t, E(VP_-)) \\
    \overline{\mathbf{\epsilon}}_{\theta}(\mathbf{x}_{t}, t, \mathbf{p}) &= 
        (1-w)\epsilon_{\theta}(\mathbf{x}_t, t) + w(\epsilon_{\theta}(\mathbf{x}_{t}, t, \mathbf{p}) + \beta \mathbf{\epsilon}_{vp}(\mathbf{x}_{t}, t))
        \label{eq:guidance}
\end{align}
where $\epsilon_{\theta}$ is Stable Diffusion's denoising U-Net and $w$ is the guidance scale. The changes to the denoising process are shown in~\cref{alg:cap}.

\begin{algorithm}[H]
\caption{Personalized image generation with ViPer}\label{alg:cap}
\begin{algorithmic}

\State $\mathbf{p} = E(p) + \beta (E(VP_+) - E(VP_-))$
\State $\mathbf{x}_T \sim \mathcal{N}(\mathbf{0},\ \mathbf{I})$
\For{$t = T , \dots , 1$}
    \State \textcolor{gray}{// Noise prediction step}
    \State ${\mathbf{\epsilon}}_{vp}(\mathbf{x}_{t}, t) = \mathbf{\epsilon}_{\theta}(\mathbf{x}_{t}, t, E(VP_+)) - \mathbf{\epsilon}_{\theta}(\mathbf{x}_{t}, t, E(VP_-))$ \qquad\qquad \textcolor{gray}{ // VP pred. noise} \quad
    \State $\overline{\mathbf{\epsilon}}_{\theta}(\mathbf{x}_{t}, t, \mathbf{p}) = 
    (1-w)\epsilon_{\theta}(\mathbf{x}_t, t) + w(\epsilon_{\theta}(\mathbf{x}_{t}, t, \mathbf{p}) + \beta {\mathbf{\epsilon}}_{vp}(\mathbf{x}_{t}, t))$ \textcolor{gray}{// Final pred. noise}

    \State \textcolor{gray}{// Sampling step}
    \State $\mathbf{z} \sim \mathcal{N}(\mathbf{0},\ \mathbf{I})$ if $t > 1$, else $\mathbf{z} = 0$
    \State $\mathbf{x}_{t-1} =\frac{1}{\alpha_t}(\mathbf{x}_{t} - \frac{1-\alpha_t}{\sqrt{1-\overline{\alpha_t}}}\overline{\epsilon}_{\theta}(\mathbf{x}_{t}, t, \mathbf{p})) + \sigma_t \mathbf{z}$
\EndFor \\
\Return $\mathbf{x}_{0}$
\end{algorithmic}
\end{algorithm}

\subsection{Proxy for evaluating personalized generations} \label{sec:method-eval}
We proposed a method to generate personalized images. To evaluate them, human evaluations are the gold standard, however, they are expensive and not scalable. In this section, we aim to introduce a proxy measure for evaluating these generations, thereby encouraging further research in this domain. 

\begin{figure}[H]
  \centering
  \includegraphics[width=1.\linewidth]{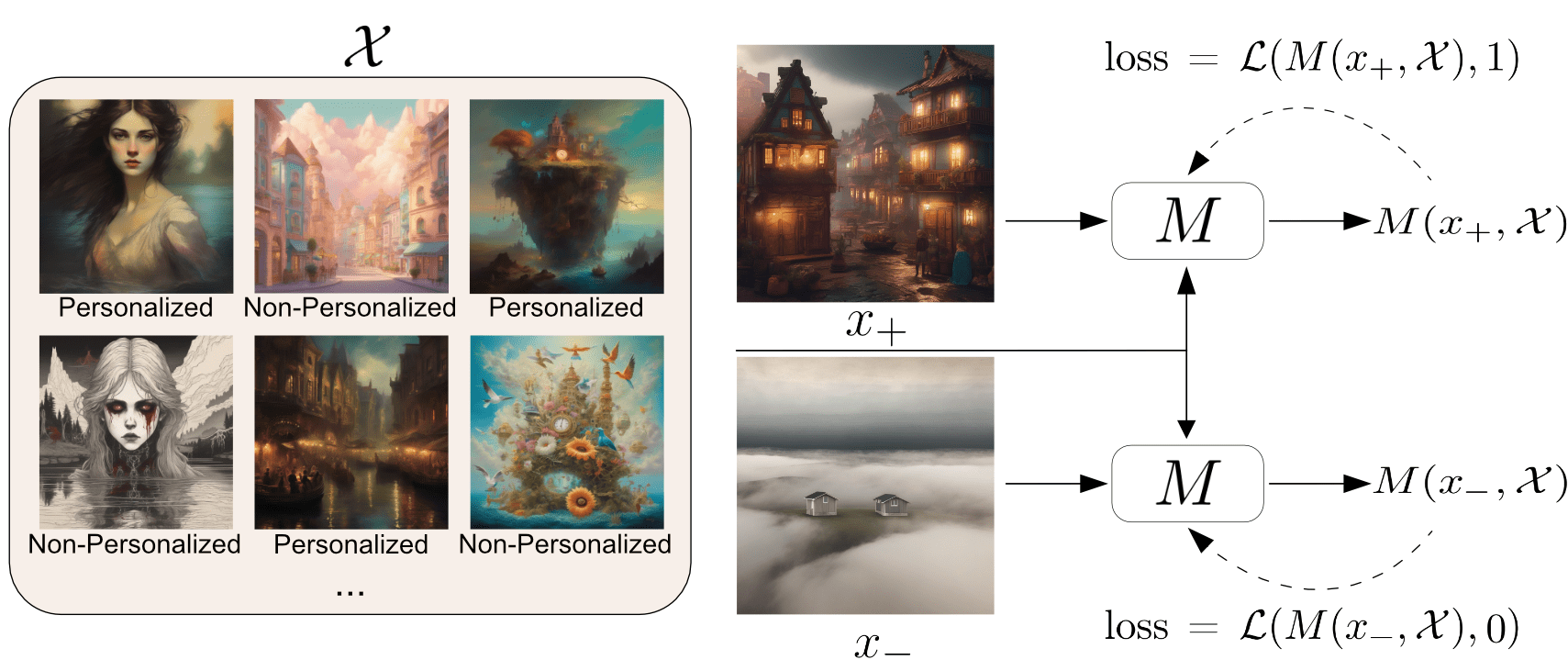}
  \caption{\textbf{Learning a proxy measure for evaluating personalized generations.} We fine-tune IDEFICS2-8b, denoted by $M$, using both the personalized/liked and non-personalized/disliked set of images, $\mathcal{X}$, of an agent. $M$ is given the set of images $\mathcal{X}$ from an individual for context and asked to predict if the user will like a given query image. We trained the model with cross-entropy loss.
  }
  \label{fig:idefics}
\end{figure}

This proxy measure would indicate how much an individual would like a given image. To obtain this proxy measure, we combine two sources of training data: one from our agents and another from the Pick-a-Pic dataset~\cite{kirstain2023pickapic}. The latter contains the binary preferences of human users on a set of generated images. For the former, we employ two approaches to generate training data. Using the agents' visual preferences, we produce a small set of personalized images for each agent with ViPer. Additionally, we generate personalized prompts with a language model and create images from them using standard Stable Diffusion XL (see \cref{sec:userevalbaselines} and \cref{sec:promptpers} for details). This is to ensure that the proxy measure will not be biased toward the generations from ViPer. Thus, for a given entity (agent or human), we either have a set of their personalized images or a set of several generated images that they like. We denote each image from this set for entity $i$ as $x^i_+$. An image that is disliked or not personalized is denoted as $x^i_-$. We denote the set of these image pairs as $\mathcal{X}^i = \{x_+^{i,j},x_-^{i,j}\}_{j=1}^k$. We use 5000 sets of image pairs to fine-tune IDEFICS2-8b, denoted by $M$, to output the probability of an individual liking a given image $x$, given a set of their liked and disliked images, $\mathcal{X}$, as $M(x, \mathcal{X}) = \Pr(x \text{ is liked by the user} \mid \mathcal{X})$.
We use the cross-entropy loss for training, $\mathcal{L}:y,\hat{y}\rightarrow \mathbb{R}$ where $y$ is the prediction and $\hat{y}$ is the label \ie, $\hat{y}\in\{0,1\}$. See~\cref{fig:idefics} for the schematics of a fine-tuning step.

\section{Evaluation and Ablations}\vspace{-2.5em}
We demonstrate that ViPer is capable of generating images that align with an individual's preferences (see~\cref{sec:evalpersuser}), and when compared to current baselines, it achieves higher levels of user satisfaction (see~\cref{sec:userevalbaselines}). \vspace{-6em}
\subsection{Evaluating the personalized generations}\label{sec:evalpersuser}\vspace{-2em}

\begin{figure}[H]
  \centering
  \includegraphics[width=\linewidth]{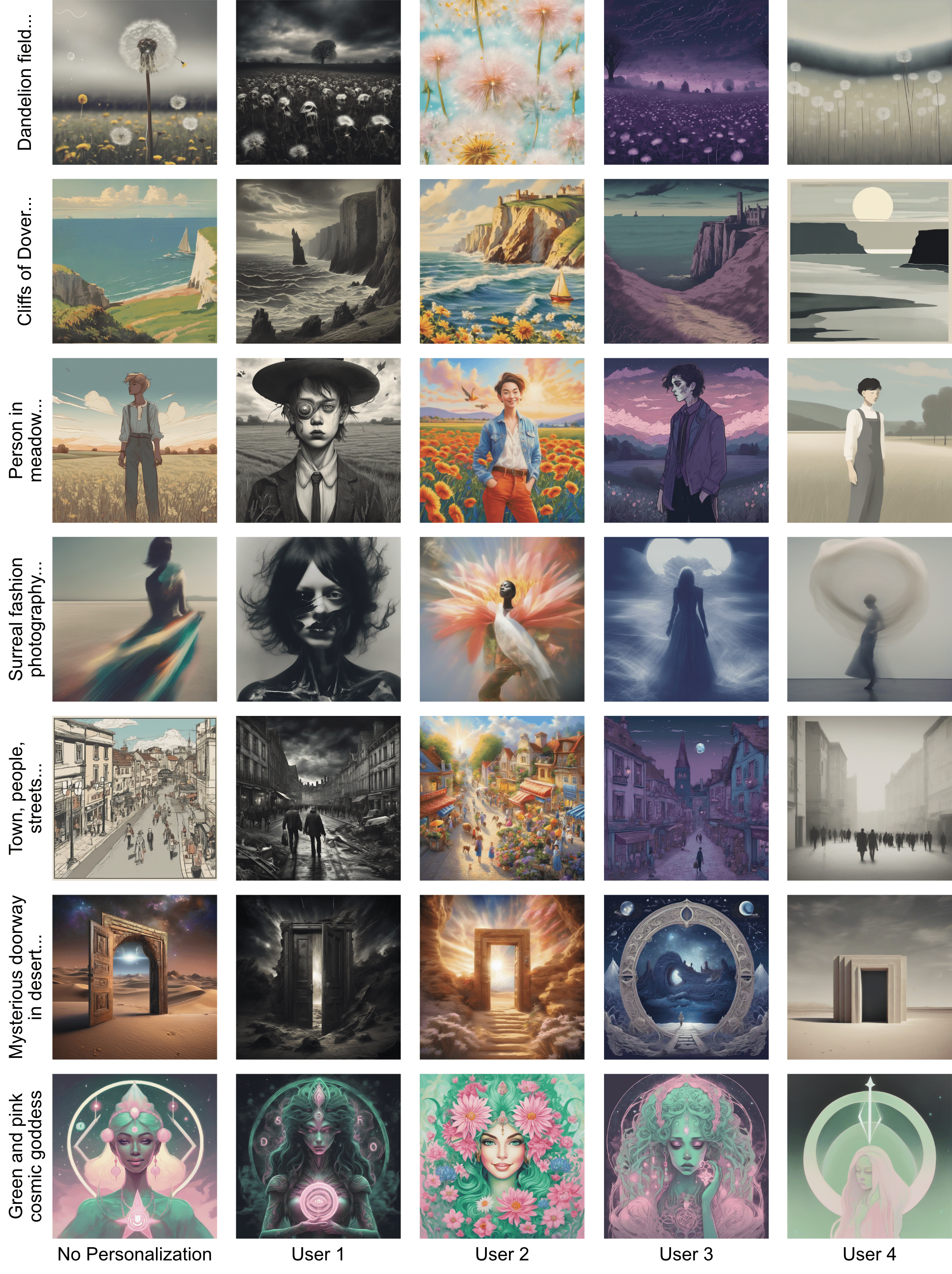}
  \caption{\textbf{Comparing the personalized generations between users.} Each row shows images generated from the same prompt shown on the left. The first column displays generations \textit{without any personalization}, while the next four columns show personalized generations for users with distinct preferences. Note that the generations are \textit{consistent with the input prompts}, and each user's preferences are reflected across prompts. Moreover, while the color palette is a dominating visual attribute and can be noticed at first glance, other visual attributes such as brush strokes, lines, vibe, etc., also have intricate effects (see last row where colors are specified through input prompt). Users' visual preferences are included in \cref{tb:VPS}.}
  \label{fig:figureonly}
\end{figure}

\textbf{Comparing generation across users.} In~\cref{fig:figureonly}, we display personalized outcomes from seven prompts for four users using ViPer. These results highlight each user's distinct art preferences, underlining that visual taste is more than just liking a single art style or a color palette; rather it's shaped by various visual attributes based on individual preferences.

\textbf{User studies.} 
We assess our approach using user tests. We invited twenty users to our study, conducting two types of tests. Each participant was first asked to comment on 8 to 20 images (flexible and dependent on the user's choice). Each user was then shown eight images generated from the same prompt, where only one of the eight was personalized for them using ViPer. This was then repeated for 15 prompts.

In the first test, the other seven images were created using standard Stable Diffusion XL, without personalization (\textit{User's Pers. vs. No Pers} in \cref{tab:usertests}). Our top-one accuracy here is 86.1\%, meaning the user's most preferred images were overwhelmingly generated from their visual preference, surpassing random chance by 6.89 times.

 In the second test, they were generated based on the visual preferences of other users (\textit{User's vs. Other Users' Pers} in \cref{tab:usertests}). The aim of this test is to showcase how ViPer goes beyond just enhancing general aesthetics and instead focuses on personalization. Our top-one accuracy here is 65.4\%, which exceeds random selection by 5.23 times. 
 
 The lower accuracy in the second test compared to the first one is because some users have similar preferences. Despite our efforts to address this for the sake of demonstration, users may still favor details in others' results that weren't explicitly mentioned in their own comments. We discuss this further in~\cref{sec:limit}. Our proxy metric's results closely correlate with actual user tests, demonstrating its alignment with user preferences. Additional ablation studies can be found in the \cref{sec:additionalex}.

\begin{table}[t]
\caption{\textbf{User study results.} In the \textit{User's Pers. vs. No Pers} test, for a given prompt, each user is shown 8 images, one of which is personalized for them and the others are without any personalization. In the \textit{User's vs. Other Users' Pers} test, each user is shown an image personalized for them and those personalized for 7 other users. In both tests, the users had to rank the images based on how much they appealed to them. This is done for 12-15 prompts per user. The last column shows the accuracy from random guessing. We report the top-1 to 4 accuracy~(\%), in columns 1 and 3 in the table below. We also show the accuracy from using our proposed proxy metric (see~\cref{sec:method-eval}). The accuracy from the user and the proxy metric are similar, indicating that the metric can be a proxy for user preferences.} 
  \label{tab:usertests}
  \centering
\adjustbox{width=0.8\textwidth}{
\begin{tabular}{c|cc|cc|c}
\toprule
\multicolumn{1}{l|}{\multirow{2}{*}{Accuracy}} & \multicolumn{2}{l|}{\textit{User's Pers. vs. No Pers}} & \multicolumn{2}{l}{\textit{User's vs. Other Users' Pers}} & \multicolumn{1}{|l}{} \\ 
\multicolumn{1}{l|}{} & Users & Proxy metric & Users & Proxy metric & Random \\ \midrule
Top-1 & 86.1 & 86.7 & 65.4 & 71.6 & 12.5 \\
Top-2 & 93.2 & 95.8 & 82.2 & 89.2 & 25.0 \\
Top-3 & 98.6 & 100.0 & 96.6 & 98.3 & 37.5 \\
Top-4 & 99.0 & 100.0 & 99.0 & 99.2 & 50.0 \\\bottomrule
\end{tabular}}
\end{table}

\begin{figure}[H]
  \centering
  \includegraphics[width=1.\linewidth]{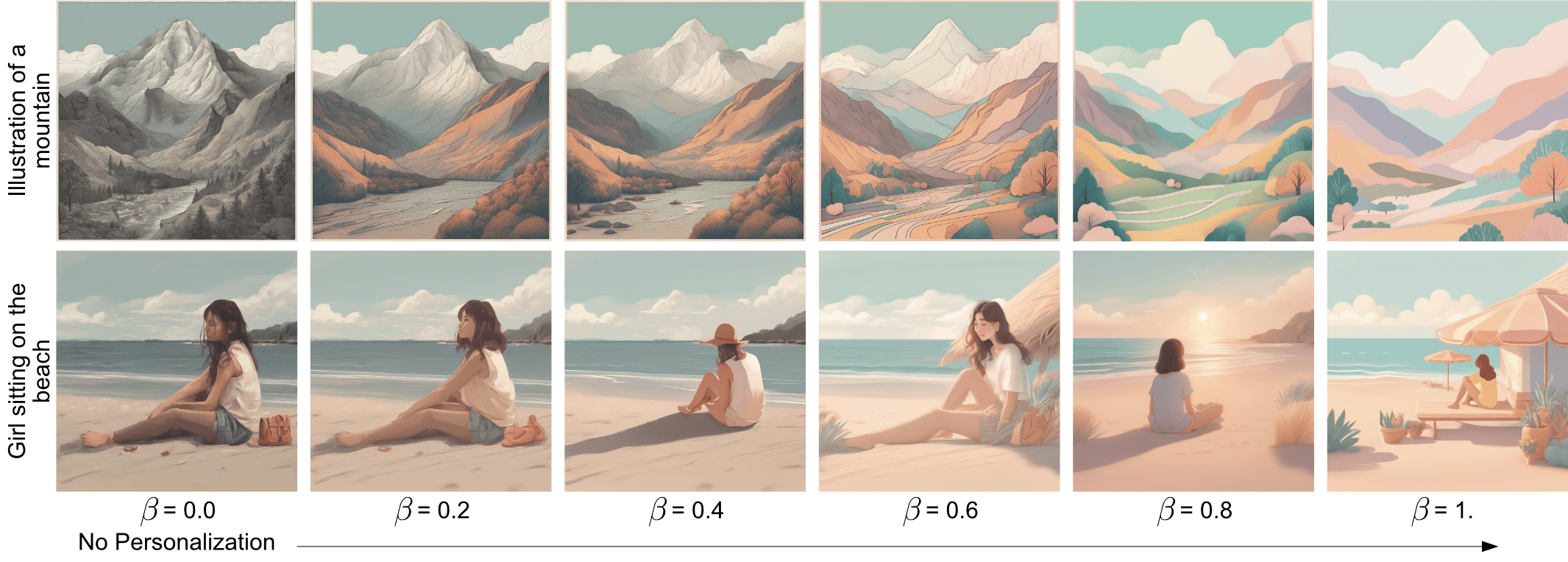}
  \caption{\textbf{Controlling the degree of personalization.} Each row displays images generated from the same prompt. The left column is generated without any personalization, \ie, $\beta = 0$ in the guidance equation in~\cref{eq:guidance}, while the next five columns increase this number by 0.2 consecutively. This user's preference for soft, simple, dreamy vibes, and pastel colors increases in intensity as $\beta$ increases.}
  \label{fig:pers_degree}
\end{figure}

\textbf{Controlling the degree of personalization.}
Our approach offers flexibility by letting users adjust the personalization level with a simple slider. ~\cref{fig:pers_degree} depicts how personalization increases for one user gradually. As $\beta$ increases from 0 to 1, results become more tailored to user preferences, departing from generic outcomes while keeping the original prompt's intent.

\textbf{Consistency with input prompts.} In our sample runs, results follow the input prompts' meaning. This is because the influence of visual preference embeddings on the final images can be limited by $\beta$, ensuring they never outweigh the input prompt embeddings. In~\cref{fig:figureonly}'s last row, we display the results of a prompt following specified art features. Despite maintaining personalization, our results adhere to the prompt's characteristics. Additional examples and experiments are provided in \cref{sec:consistency}.

\textbf{Evaluation through identifiability tests.} We conduct two experiments to make sure that there is a significant connection between the agents' preferences and their comments so that $VPE$ is able to extract preferences correctly from user comments.

In the first experiment, which we denote as "Author test" in \cref{tab:table2}, we aim to identify the author of a comment given the visual preferences of five agents. We conduct this using 100 random agents, each appearing at least twice as the author. GPT-4 ranks agents by likelihood of being the author and we measure accuracy by whether the true author is among the top $k$ predictions.

The second experiment, which we call "VP vs. VP" in \cref{tab:table2}, aims to assess how well an agent's set of comments maintains their original visual preference. To do this, we first use GPT-4 to reconstruct the visual preferences of 50 random agents from all of their comments. Then for each agent, we consider the agent's reconstructed visual preference (which was generated from the comments), the agent's original visual preference (which the comments were generated from), and four random agents' original visual preferences. We instruct GPT-4 to rank the reconstructed visual preference with the original visual preferences based on the likelihood of both belonging to the same agent.

In both experiments, the top-one accuracy exceeds random selection significantly. The accuracy of the second test, VP vs. VP, is overall higher than the first test. This is because the reconstructed visual preferences contain more information than one comment, making the second test easier in general. These results are proof of alignment between our simulated agents' preferences and their comments, making $VPE$'s training data reliable as a source to extract visual preference from comments.

\begin{table}[t]
  \centering
\caption{In the Author test, GPT-4 is instructed to identify the author of a comment among five given agents' visual preferences by ranking them based on the likelihood of being the author. For the VP vs. VP test, we first reconstruct the visual preferences of agents from their comments with GPT-4, and then we instruct it to identify an agent given their reconstructed visual preference and five ground truth visual preferences, with only one belonging to the user. We perform variations of this test using both positive and negative visual preferences, denoted by $VP$, and each separately, denoted by $VP_+$ and $VP_-$. The goal of these tests is to ensure alignment between comments and visual preferences so that we're sure $VPE$ which is tuned on this data, can accurately extract visual preferences from comments. Note that the top-k results (\%) exceed the random chance of agents being chosen by the language model, demonstrating the link between agents' comments and their visual preferences. Additionally, it's worth noting that in the second test, GPT-4 performs more accurately when given either the entire visual preference or its positive aspect.} 
\begin{tabular}{c|c|c|c|c|c}
\toprule
\multirow{2}{*}{Accuracy} & \multirow{2}{*}{Author test} & \multicolumn{3}{c|}{VP vs. VP test}                                 & \multirow{2}{*}{Random} \\
                          &                              & $VP$                        & $VP_+$               & $VP_-$ &                         \\ \midrule
Top-1                     & 83.4                         & \multicolumn{1}{c|}{89.9} & \multicolumn{1}{c|}{92.4} & 45.8        & 20.0                      \\
Top-2                     & 89.1                         & \multicolumn{1}{c|}{98.1} & \multicolumn{1}{c|}{96.2} & 72.1        & 40.0                      \\
Top-3                     & 91.2                         & \multicolumn{1}{c|}{100.0}  & \multicolumn{1}{c|}{96.4} & 88.3        & 60.0                      \\ \bottomrule
\end{tabular}\label{tab:table2}
\end{table}

\subsection{Comparison with baselines} \label{sec:userevalbaselines}
In this subsection, we compare our approach to existing baselines. We'll discuss the strengths and weaknesses of these methods and explain the advantages of our approach over them.
We evaluate the following baselines and use Stable Diffusion 2 for the experiments:
\begin{itemize}[leftmargin=*,label={}]
    \setlength\itemsep{0.3pt}
    \item \textit{No personalization}: These generations are produced with the input prompts using standard Stable Diffusion.
    \item \textit{ZO-RankSGD}~\cite{tang2024zerothorder}: Users are required to rank several images in multiple iterations for each prompt. This algorithm converges to a stationary point based on the feedback received. This approach helps us compare our approach to ranking.
    \item \textit{FABRIC}~\cite{vonrütte2023fabric}: For each prompt, users are asked to like/dislike the generated images in multiple steps. The diffusion process is conditioned on this set of liked and disliked feedback images. This method helps us understand which one—binary reactions or comments—is more effective in grasping user preferences.
    \item \textit{Textual Inversion}~\cite{gal2022image}: Users provide two sets of liked and disliked images. Textual Inversion generates special tokens for each. Tokens are incorporated into positive and negative prompts. This approach helps determine whether asking a user to provide their preferred and disliked images, or comment on images is better at capturing user preferences.
    \item \textit{Fine-tuning Stable Diffusion}~\cite{gal2022image}: We fine-tune Stable Diffusion with one user's $VP$, using the CLIP similarity score to align Stable Diffusion-generated outputs for each user with their $VP$ to serve as a reward model \cite{xu2023imagereward, clark2023directly}. The key distinction from ViPer is in the incorporation of visual preferences into Stable Diffusion. This approach demonstrates the effectiveness of using $VP$ embeddings as shown in \cref{eq:embeddingaddtion} and \cref{eq:guidance}.
    \item \textit{Prompt Personalization}: We instruct a language model to directly personalize prompts, based on the user's visual preferences extracted from their comments by $VPE$. The modified prompt is given to standard Stable Diffusion to generate personalized images. The main difference with ViPer lies in how visual preferences are incorporated into Stable Diffusion. Thus, this method demonstrates the efficacy of using $VP$ embeddings in \cref{eq:embeddingaddtion} and \cref{eq:guidance}.
\end{itemize}

\begin{figure}[t]
  \centering
  \includegraphics[width=1.\linewidth]{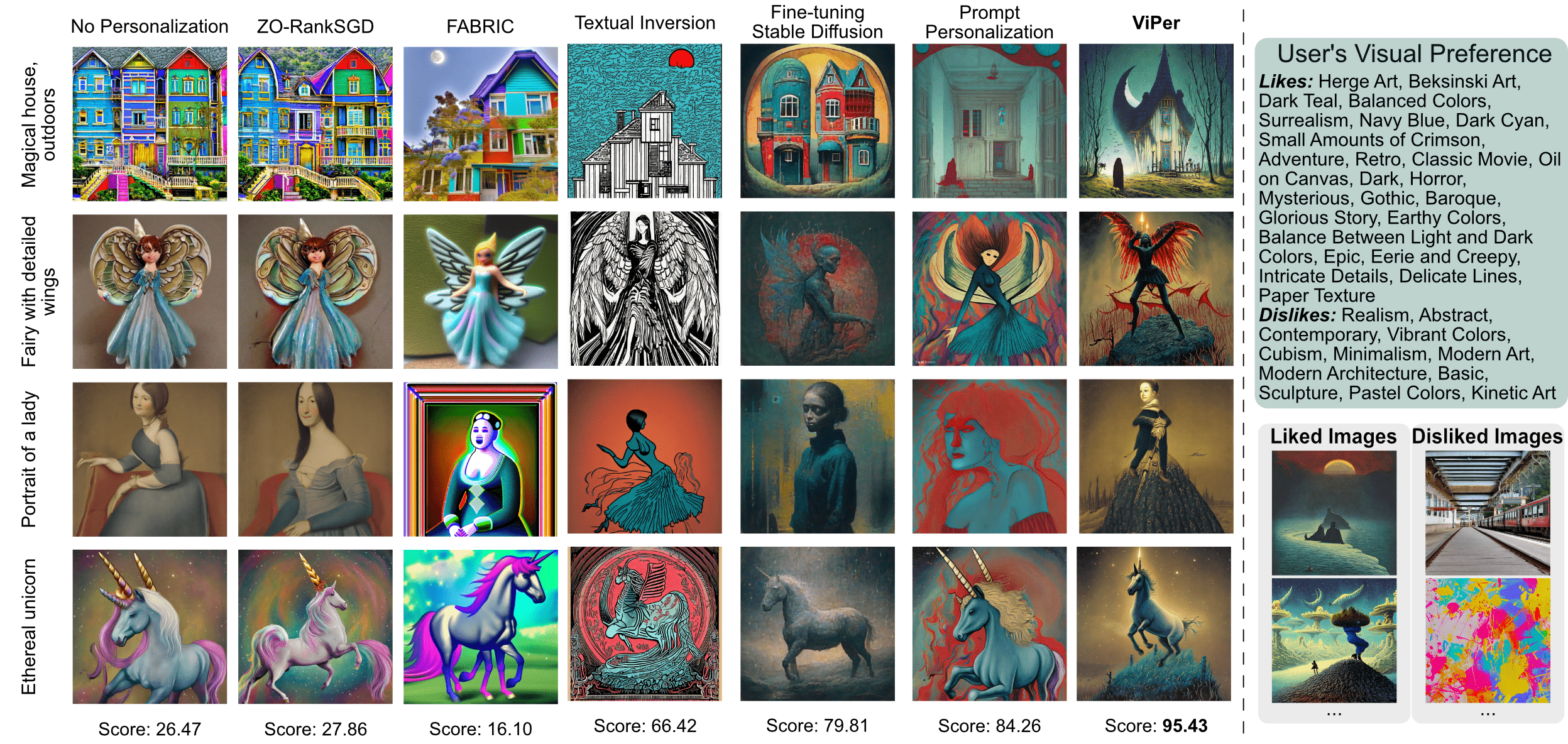}
  \caption{ 
  \textbf{Left: Comparing the generations from ViPer versus the baselines for a single user.} Each row displays images generated from the same prompt (except Prompt Personalization which manipulates the input prompt). The leftmost column is generated without any personalization, while the next five columns shows the generations from different methods.
  Note how ZO-RankSGD's generations are similar to the generations without any personalization. FABRIC tends to generate images that are more realistic or saturated, which is not aligned with this user's preferences. Textual Inversion's learned patterns does not align with the user's taste. Fine-tuning Stable Diffusion results in low-quality generations but there is some degree of personalization. Prompt Personalization generates images that are close to the user's preferences, however, they are not always consistent with the input prompt (\eg, the first image in the fifth column). Finally, ViPer generates images that are both aligned with the user's preferences and follow the input prompts. The user consistently preferred ViPer's results. Our proxy metric's average score for each method is presented at the bottom of each column. \textbf{Right: User's preference info.} A user's visual preference is extracted from their comments. Some of the user's liked images and disliked images that were used for our metric's context and Textual Inversion's input images are shown below. Note how there is an alignment between the user's visual preference, set of liked and disliked images, and ViPer's results. The alignment is significantly less obvious in FABRIC and ZO-RankSGD.
  }
  \label{fig:compare_eval}
\end{figure}

~\cref{fig:compare_eval} depicts a comparison of personalized generations with these baselines versus ViPer, for a single user.
Textual Inversion struggles when the user's selected images are too diverse. Furthermore, users are not able to provide feedback on their reasons for liking/disliking the input images. This can be seen in column 4 in~\cref{fig:compare_eval}, where the results contain textures and certain color palettes that are too extreme for the user's taste. It is also challenging for users to provide a set of images that completely represents their preferences.

Prompt Personalization, which is dependent on a language model, may lead to lengthy prompts, parts of visual preferences being disregarded, and can adopt a conversational tone, despite explicit instructions. Additionally, the personalized prompts may also fail to include key concepts from the original prompts (examples in \cref{sec:consistency}). The resulting generations can be seen in the fifth column of~\cref{fig:compare_eval}. The results are  aligned with the user's visual preference but overlook the user's favored attributes like "Retro", "Creepy", and "Horror" due to language models' inclination towards generating lengthy outputs. Iterative prompt refinement helps, but the computational overhead of using a language model remains a challenge. We provide further examples of these shortcomings in \cref{sec:consistency}.

Fine-tuning Stable Diffusion results in personalized but low-quality outputs. Additionally, it is an expensive and time-consuming method because each user needs to fine-tune Stable Diffusion with their $VP$.

In our experiments, FABRIC consistently converged towards a more realistic and vibrant style of images. This can be seen in the third column in~\cref{fig:compare_eval}, despite the user's dislike for "Realism" and "Vibrant Colors". ZO-RankSGD does not deviate much from the unpersonalized results.

Lastly, we use our proxy metric to calculate a personalization score for each of the methods in ~\cref{fig:compare_eval}. ViPer has the highest score, followed by Prompt Personalization and Fine-tuning Stable Diffusion. This aligns well with how the users ranked the images based on their preferences (we present the user's ranking in \cref{fig:ranking}).

\section{Conclusion and Limitations}\label{sec:limit}
We introduced ViPer, a method for personalizing generative models. It extracts and incorporate individuals' generic preferences into Stable Diffusion. Moreover, we introduced a metric to automatically assess personalization, reducing the need for costly human evaluations of generated images. Our approach is flexible and scalable and can consider a wide range of preferences. We briefly discuss the limitations and potential future work:

\textit{Exploring other ways of extracting user's preferences.} Our method makes use of user's comments on a given set of images. As discussed in~\cref{sec:method-vp}, this is more expressive than \eg, ranking images, however, articulating these preferences can still be difficult for some users. An ideal approach could involve accessing human brain signals directly or scalable proxies of it. Another useful angle is engineering the set of commentary images to maximally reveal the users' preferences with least interactions. 

\textit{Set of visual attributes.} We pre-defined a set of structred visual attributes, but despite our best efforts, it may not cover everything. To address this limitation, we explored using more powerful language models like GPT-4, eliminating the need for pre-defined attributes and allowing more flexibility in the interpretation of comments. However, these models are not open-sourced, and using them at a large scale can be expensive.

\textit{Stable Diffusion limitations.} Our current implementation relies on Stable Diffusion's text encoder, which has limitations such as a 77-token restriction and sensitivity to word order, sometimes leading to some preferred attributes being overlooked. Randomization techniques can partially address this issue, but using an encoder without these limitations would be beneficial. Furthermore, if Stable Diffusion cannot grasp a visual attribute due to its training data distribution, it cannot include it in its results.

\textit{Exploring other ways of adapting Stable Diffusion.} We attempted to fine-tune the Stable Diffusion by optimizing our proposed proxy introduced in~\cref{sec:method-eval}. However, the generations suffered in quality. Exploring alternative reward tuning strategies is an interesting future direction.

{\small
\bibliographystyle{plainnat}
\bibliography{main}
}

\clearpage
\setcounter{section}{0}

\appendix

\section*{\LARGE Appendix}
\startcontents[appendices]
\printcontents[appendices]{l}{1}{\setcounter{tocdepth}{2}}
\newpage

\section{Experimental Setup}\label{sec:exsetup}
\subsection{Visuals of Generated Images Used for Comments}
\begin{figure}[H]
  \centering
  \includegraphics[width=\linewidth]{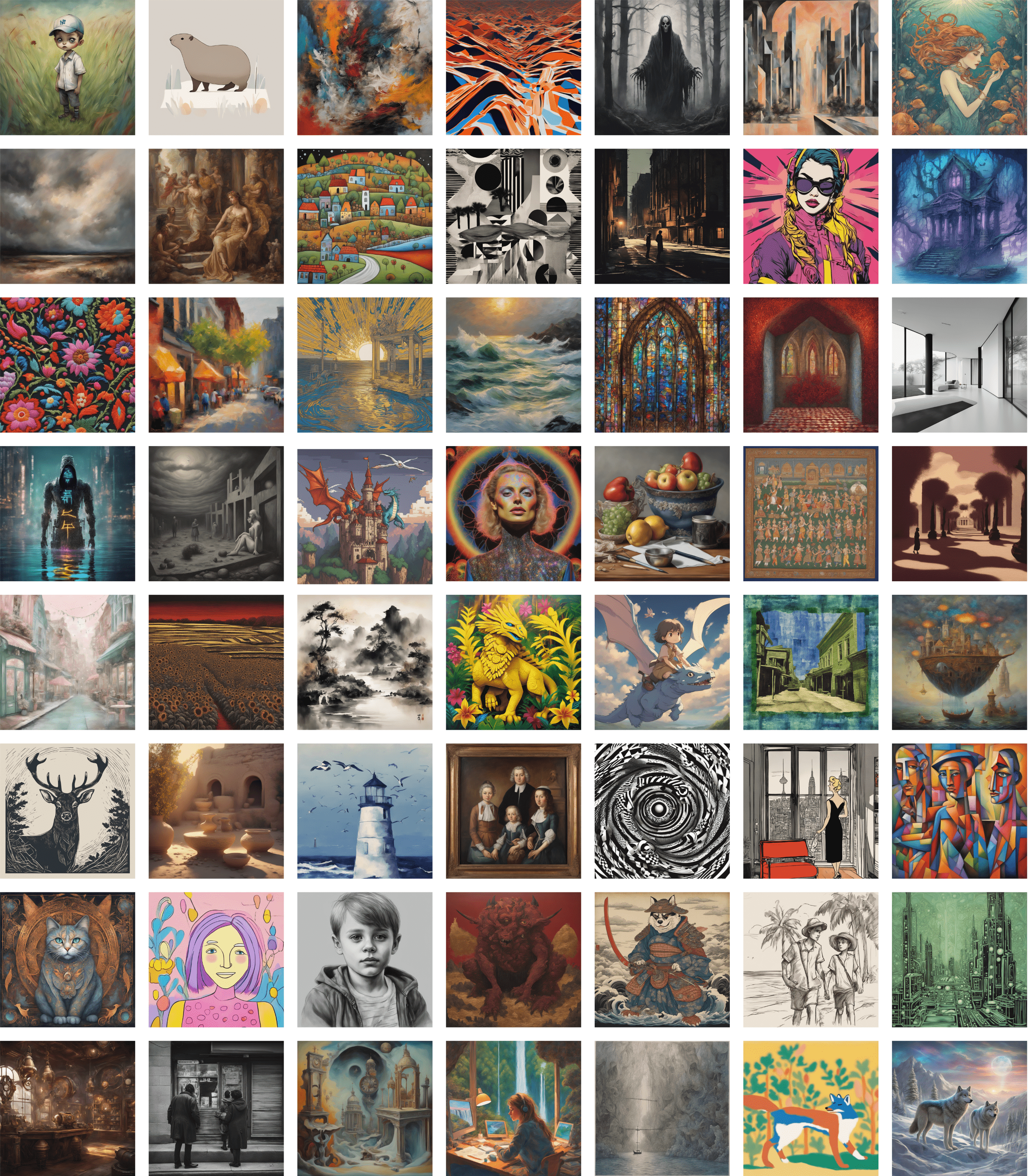}
  \caption{
  Query image set: We generate around 50 images with different art styles and concepts. Individuals are encouraged to choose a subset of these images to comment on.
  }
  \label{fig:commentimages}
\end{figure}
\label{sec:commentimages}
We employ Stable Diffusion XL \cite{rombach2022highresolution} to generate around 50 images spanning diverse art styles and concepts. To ensure diversity, we first generated approximately 2000 visual attributes using the same method we used for creating our agents' visual preferences. Next, we used multiple language models, such as Llama 3 \cite{llama3modelcard} and GPT-4 \cite{openai2024gpt4}, to generate 50 concepts. We then concatenated the visual attributes with these concepts and selected the top 50 unique images generated from them based on CLIP similarity. Finally, we manually included some images that seemed missing from the final set. While acknowledging that art cannot be fully represented by just 50 images, we opted for this smaller set to spare individuals from the burden of choosing from an extensive selection. However, recognizing the limitations in capturing all aspects of art with this limited set, we ask individuals to provide their own images to comment on if our set fails to evoke strong reactions. We show the generated images in \cref{fig:commentimages}.

\subsection{Prompt Templates} \label{sec:promptemp}
We use the following prompt template to generate comments using GPT4.
\begin{lstlisting}
{Image attached.} Task: Write a short and simple Comment on the provided artwork.
Comment Requirements:
1. Express an opinion (neutral, extremely negative, negative, positive, conflicted) about the artwork.
2. Make the comment short, efficient, and reflective of the specified visual attributes liked or disliked.
3. The comment shouldn't specify the given artwork's features directly.
4. The comment shouldn't hallucinate about the artwork's features or the given preference.
5. The comment should be in first person POV.
Provided Information:
Liked visual attributes: {VP+}
Disliked visual attributes: {VP-}
Instruction:
From the perspective of the provided personality and considering the liked and disliked visual attributes, write a brief comment on the given artwork. Your comment should succinctly express an opinion that reflects these preferences while giving insight into the individual's taste in art. Only use simple words and avoid poetic or complex structures.
\end{lstlisting}

The prompt template for fine-tuning $VPE$ is as follows:

\begin{lstlisting}
I will provide a set of artworks along with accompanying comments from a person. Analyze these artworks and the comments on them and identify artistic features such as present or mentioned colors, style, composition, mood, medium, texture, brushwork, lighting, shadow effects, perspective, and other noteworthy elements.
Your task is to extract the artistic features the person likes and dislikes based on both the artworks' features and the person's comments. Focus solely on artistic aspects and refrain from considering subject matter.
If the person expresses a preference for a specific aspect without clearly stating its category (e.g., appreciating the colors without specifying which colors), identify these specific features from the images directly to make the person's preference understandable without needing to see the artwork.
Your output should consist of two concise lists of keywords: one listing the specific art features the person likes and another listing the specific features they dislike (specified in keyword format without using sentences).
Here are the images and their corresponding comments:
{Image 1} {Comment 1}
{Image 2} {Comment 2}
...
{Image k} {Comment k}
\end{lstlisting}

\subsection{Visual Attributes} \label{sec:visattr}
\begin{figure}[h!]
  \centering
  \includegraphics[width=1.\linewidth]{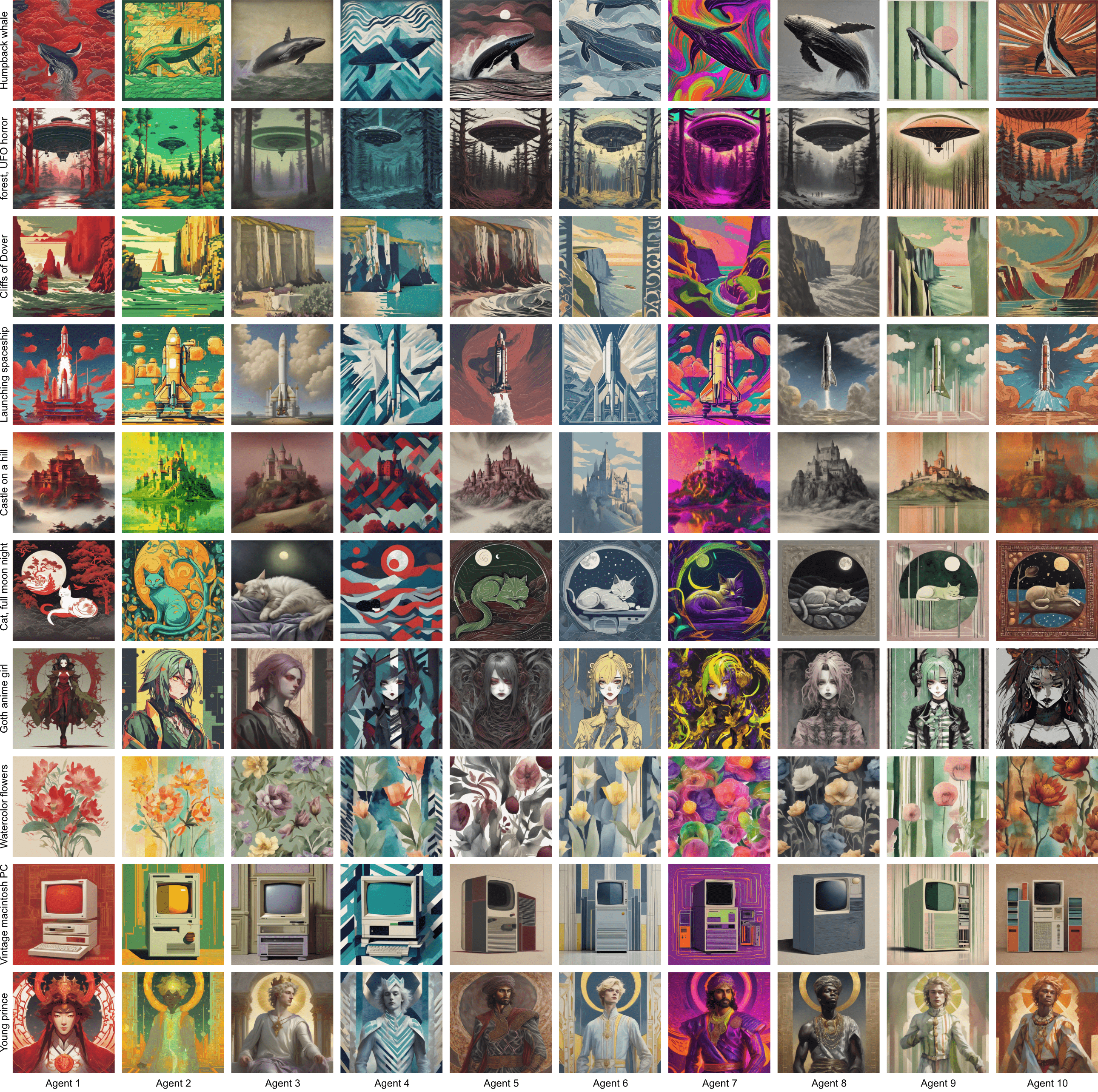}
  \caption{
  The personalized generations of 10 agents, generated with 10 different prompts from the same seed, are presented here. These agents' visual preferences differ not only in color palette and style but also in various attributes such as texture, pattern, vibe, etc. Note how the preference pattern of each agent's results remains consistent, with some being more extreme than others.
  }
  \label{fig:agentimgs}
\end{figure}

The full set of visual attributes is presented on page \pageref{visat}. In this set, we tried to consider the main features that describe and affect the individual's perception of image. Some categories include subcategories that contain similar visual attributes. We randomly select several of these attributes for each agent, prioritizing more popular categories such as style, color palette, composition, mood, and medium over movement, depth, lines, etc. This prioritization is done by giving a larger weight to popular categories that we observed mentioned frequently in Lexica's\footnote{lexica.art} prompts. The results of 10 sample agents are shown in \cref{fig:agentimgs}.

\subsection{Experiments with Other Models}\label{sec:vlm}
\begin{figure}[H]
   \centering
   \begin{minipage}[t]{0.49\linewidth}
     \centering
     \includegraphics[width=\linewidth]{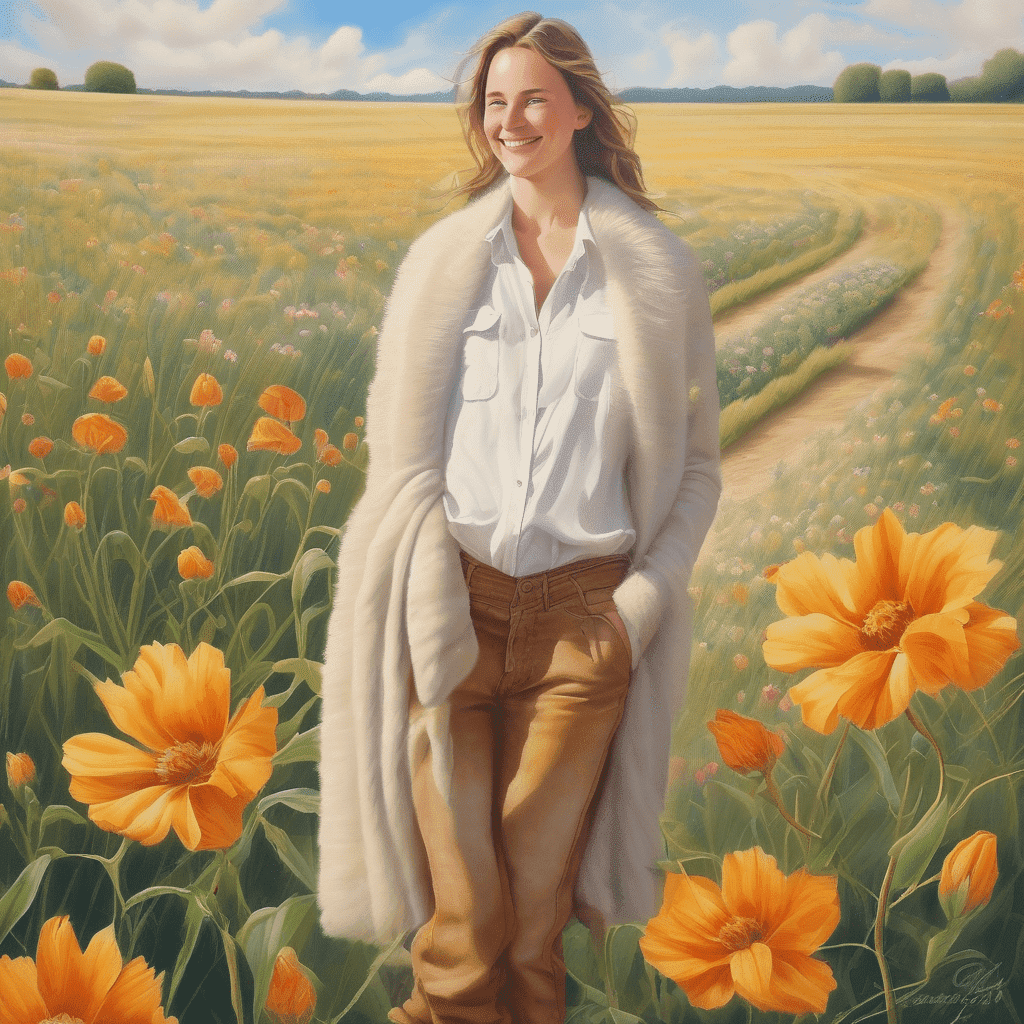}
     \captionsetup[subfigure]{width=0.9\linewidth}
     \subcaption{\textbf{IDEFICS1-80b}: I'm not really feeling this artwork. It's too busy and vibrant for my taste. I prefer minimalist and abstract pieces with subtlety and open spaces.\\\textbf{LLaVA}: I find the artwork to be unsettling and creepy, which I appreciate. The melancholic atmosphere and psychological depth and extreme details make it a thought-provoking piece.\\\textbf{InstructBLIP}: I like the artwork because it has a dark and eerie feel to it. It also has a lot of symbolism and meaning behind it, which is why I like it so much.}
     \label{fig:short-a}
   \end{minipage}
   \hfill
   \begin{minipage}[t]{0.49\linewidth}
     \centering
     \includegraphics[width=\linewidth]{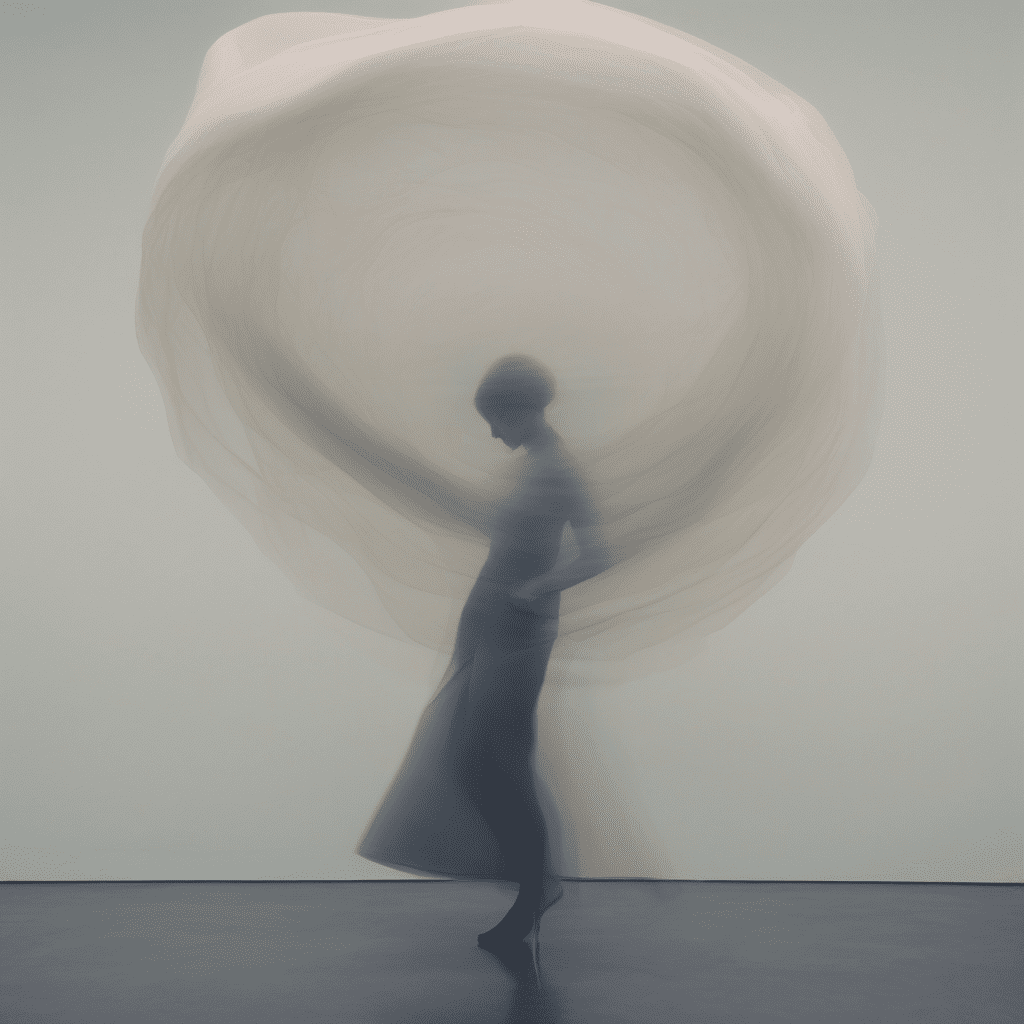}
     \captionsetup[subfigure]{width=0.9\linewidth}
     \subcaption{\textbf{IDEFICS1-80b}: I'm not sure about this artwork. It seems a bit too dark and gloomy for my taste. I prefer art that is more uplifting and positive, with vibrant colors and details.\\\textbf{LLaVA}: The sketch lines and dark themes add to the eerie vibe of the painting. The overall mood remains melancholic and psychologically deep. \\\textbf{InstructBLIP}: the painting of a woman in a field of flowers is not something that I would like to look at.}
     \label{fig:short-b}
   \end{minipage}
 \caption{The comments are generated given an image and an individual's visual preference. We present some shortcomings we witnessed for different individuals. In \cref{fig:short-a}, neither of the generated comments follows the input image's content. They discard its minimal and contemporary style. In \cref{fig:short-b}, IDEFICS1-80b and LLaVA's results do not follow the input image's content, discarding its happy vibe, while InstructBLIP ignores the given visual preference of a user who loved happy and colorful images depicting flowers.}
   \label{fig:short}
\end{figure}

We use GPT4 to generate comments since our initial experiments with the open-source vision-language models, such as IDEFICS2-8b \cite{laurençon2023obelics, laurençon2024matters}, IDEFICS1-80b \cite{laurençon2023obelics}, LLaVA \cite{liu2023visual, liu2023improved}, and InstructBLIP \cite{dai2023instructblip}, hallucinated more often than not. We show some samples in \cref{fig:short}. We input the image, and the individual's preferences, $VP_+$ and $VP_-$, to the model with the same comment generation template as the one shown in~\cref{sec:promptemp}.

The sample results of these vision-language models shown in \cref{fig:short} mostly discard the input image's semantics, vibe, and details, and sometimes even discard the given visual preference. In some extreme cases, we witnessed that LLaVA and InstructBLIP would write captions for the images rather than comments, or simply repeat some of the attributes of the given visual preferences without further context.

\section{User and Proxy Metric Alignment}\label{sec:humanproxy}
\begin{figure}[tb]
  \centering
  \includegraphics[width=\linewidth]{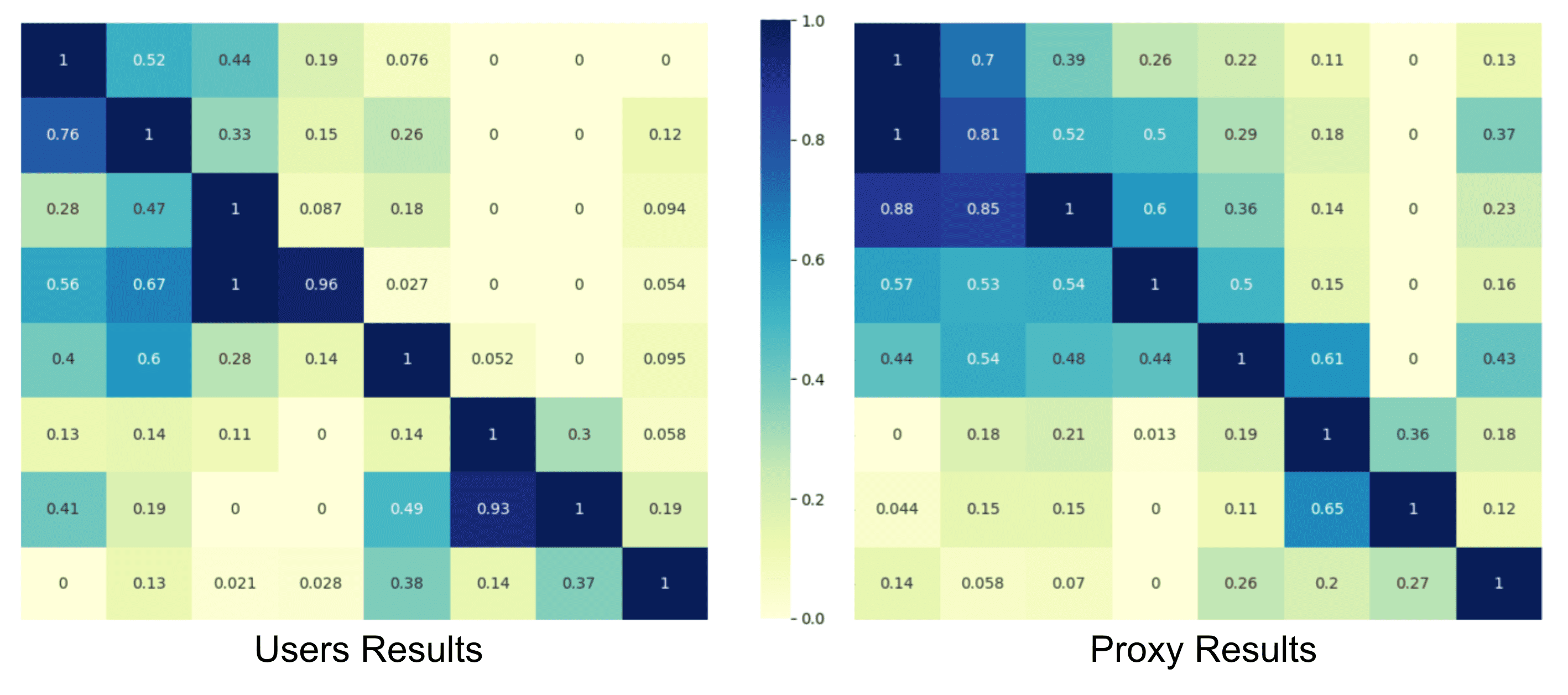}
  \caption{
  On the left, we display the results of users, and on the right, we display proxy $M$'s results from the \textit{User's vs. Other Users' Pers} test for eight users. By providing each user's preferences (liked and disliked images) to $M$, we simulate their behavior. Images generated from 15 distinct prompts personalized for the eight users are organized within each matrix according to real user rankings or $M$'s predictions. A score is assigned to indicate the preference of user $i$ towards user $j$'s images, with higher scores reflecting greater preference. The scores are averaged over prompts and the matrices are normalized for analysis. Note how both users and $M$ preferred the personalized image on average in most cases. It's crucial to understand that attaining 100\% alignment between users and $M$ is impossible due to the complexity of users' preferences, which aren't solely influenced by personalization factors. These variations are intrinsic to Stable Diffusion rather than personalization itself. $M$ is limited in its ability to accommodate these nuances as it's tuned solely to recognize users' preference patterns based on their liked and disliked images.
  }
  \label{fig:metric}
\end{figure}
To demonstrate the alignment between users and the proxy metric, we closely examine the \textit{User's vs. Other Users' Pers} test for eight users. We emulate each user's behavior with our metric $M$ by providing $M$ with their liked and disliked images. In \cref{fig:metric}, we present the confusion matrices of the user test alongside the metric's simulation of the exact same test. Within each confusion matrix, images generated from 15 distinct prompts personalized for the eight users are arranged based on real user rankings, or $M$'s predictions. Rank 1 indicates that the user (or $M$) preferred that image over the others, while Rank 8 indicates the user's (or $M$'s) least liked image. In most cases, users only ranked from 1 to 4, as they felt indifferent toward the other images present in the test for that prompt. The score in the $i$th row and $j$th column indicates how much user $i$ likes the images of user $j$: for each row $i$, a score of $\frac{1}{r}$ is added to the element $i,j$ if user $i$ ranks an image of user $j$ in the $r$th place (or if $M$ ranks it for them). These scores are then averaged over the prompts, and each row of the matrices is normalized. Note how the users' results are aligned with $M$'s predictions. On average, both users and $M$ ranked their personalized images as the most preferred. It is important to note that achieving 100\% alignment between users and $M$ is unattainable. This is due to the nuanced nature of users' choices, which are not solely dependent on personalization factors. For instance, one user selected an image that deviated significantly from their indicated preference. Their rationale for not choosing the personalized image was that "\textit{the person in this image is too thin, like she's starving}". These deviations are inherent to Stable Diffusion itself rather than the concept of personalization. $M$ is unable to account for such deviations as it is solely tuned to discern users' preference patterns from their favored and disliked images.

\begin{figure}[h!]
  \centering
  \includegraphics[width=\linewidth]{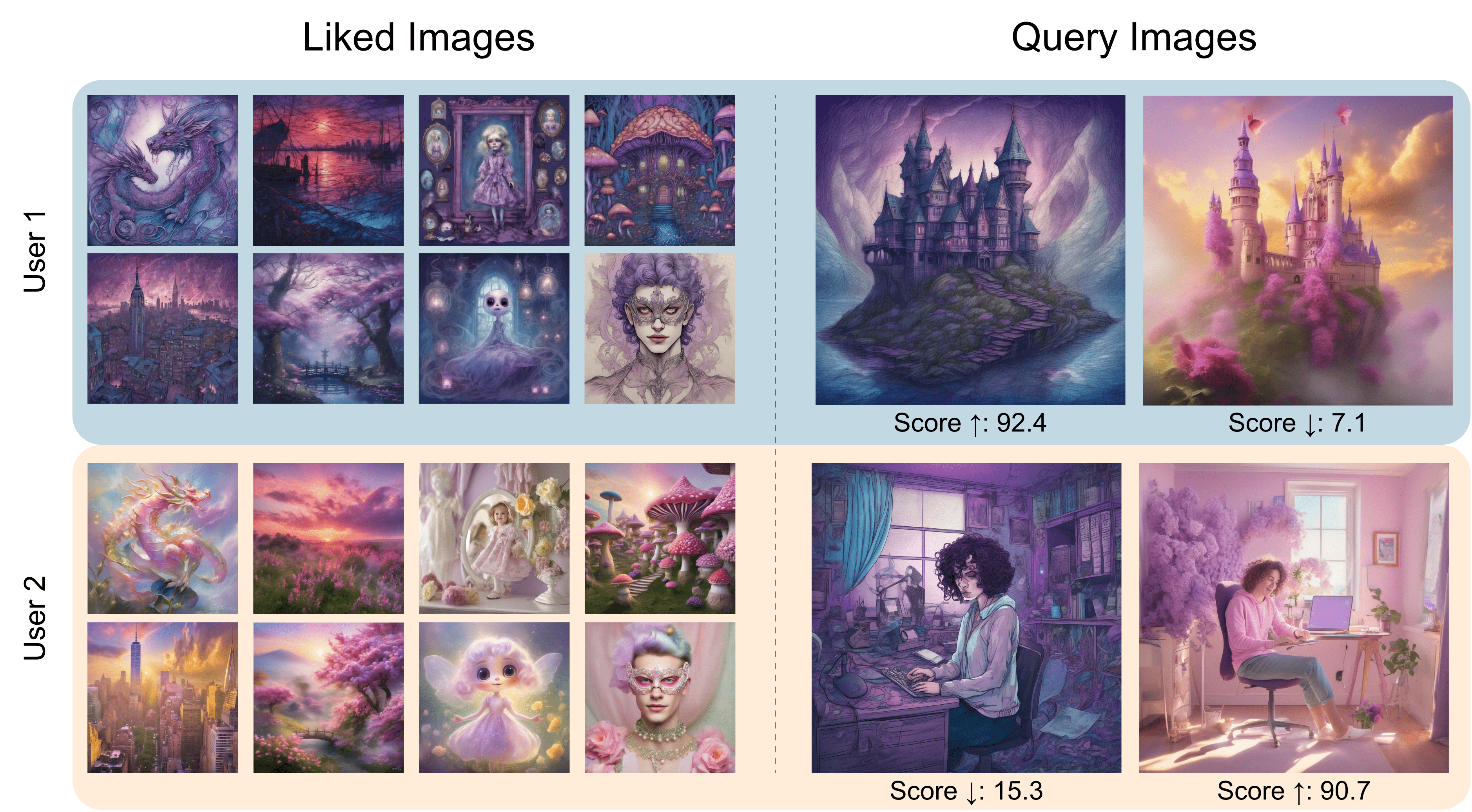}
  \caption{
  Each user is represented to $M$ by the set of images they like. Notice how, given this set of images, $M$ manages to output a higher score to query images that share the same preferred features and a lower score to those that do not.
  }
  \label{fig:scores}
\end{figure}

In \cref{fig:scores}, we demonstrate how a user is represented to $M$ by the set of images they like (disliked images are not shown for the sake of simplicity). Notice how $M$ assigns a significantly higher score to query images that share the same visual pattern and vibe as the liked images for each user.


\section{Additional Experiments}\label{sec:additionalex}

\subsection{Large-scale Automated Evaluation} \label{sec:automatedeval}
To compare the performance of ViPer with other personalization methods, we conducted multiple tests on images personalized for 1000 agents using both Prompt Personalization and ViPer. We use $M$, CLIP similarity, and PickScore \cite{kirstain2023pickapic} to calculate accuracy.\\\textbf{Personalized vs. Non-Personalized Comparison} In this experiment, we utilize the proxy metrics to distinguish between agents' personalized images and random non-personalized images. We anticipate these proxies assigning higher scores to personalized images compared to non-personalized ones. Accuracy is measured as the ratio of correct comparisons to the total number of comparisons. To conduct this test with $M$, for each agent, we input their set of personalized images as liked images and random non-personalized images as disliked images to act as the preference signal. When using CLIP and PickScore, the preference signal becomes the $VP$ text itself. Subsequently, for each agent, we consider two query images: one personalized and one non-personalized. The proxy metrics are then run on all agents, and alignment between preference signals and query images is calculated. The results are outlined in \cref{tab:table2}. It's notable that ViPer consistently outperforms Prompt Personalization across all three metrics. Despite CLIP and PickScore showing lower accuracy as they are not specifically fine-tuned for this task, their evaluations largely align with $M$'s.\\\textbf{Prompt-Image Alignment} We evaluate the alignment of base prompts and personalized images to determine which personalization method prioritizes results following the users' input prompts. For each agent, we consider two query images, both personalized. Then, we run CLIP similarity and PickScore on the initial input prompt and the queries. We anticipate that the score will be higher for the personalized image generated from the input prompt compared to the one that is not generated from that input prompt. The results can be seen in \cref{tab:table3}. Note how, once again, ViPer outperforms Prompt Personalization.

\begin{table}[h!]
\centering
\caption{We use three proxy metrics to differentiate between personalized and non-personalized images of 1000 agent, expecting higher scores for personalized ones. Accuracy is the ratio of correct comparisons to total comparisons. For each agent, their personalized images are compared to random non-personalized ones as the preference signal. With CLIP and PickScore, the preference signal is the VP text. Two query images, one personalized and one non-personalized, are considered per agent. See how ViPer consistently outperforms Prompt Personalization across all metrics. While CLIP and PickScore, not fine-tuned for this task, show lower accuracy, their evaluations largely align with $M$'s.}
\label{tab:table2}
\begin{tabular}[t]{@{}l|c|c|c@{}}
\toprule
&CLIP&PickScore&$M$\\
\midrule
Prompt Personalization&69.5\%&62.4\%&83.9\%\\ \midrule
ViPer&73.3\%&67.0\%&\textbf{89.1}\%\\
\bottomrule
\end{tabular}
\end{table}

\begin{table}[h!]
\centering
\caption{We utilize CLIP and PickScore to assess the alignment between personalized images and input prompts, anticipating higher scores for images generated from the input prompt compared to those that are not. Accuracy is measured by the ratio of correct comparisons to total comparisons. Notice how ViPer consistently outperforms Prompt Personalization across both metrics.}
\label{tab:table3}
\begin{tabular}[t]{@{}l|c|c@{}}
\toprule
&CLIP&PickScore\\
\midrule
Prompt Personalization&83.5\%&79.6\%\\ \midrule
ViPer&90.3\%&87.8\%\\
\bottomrule
\end{tabular}
\end{table}

\subsection{Effect of Number of Comments}

\begin{figure}[t]
  \centering
  \includegraphics[width=0.8\linewidth]{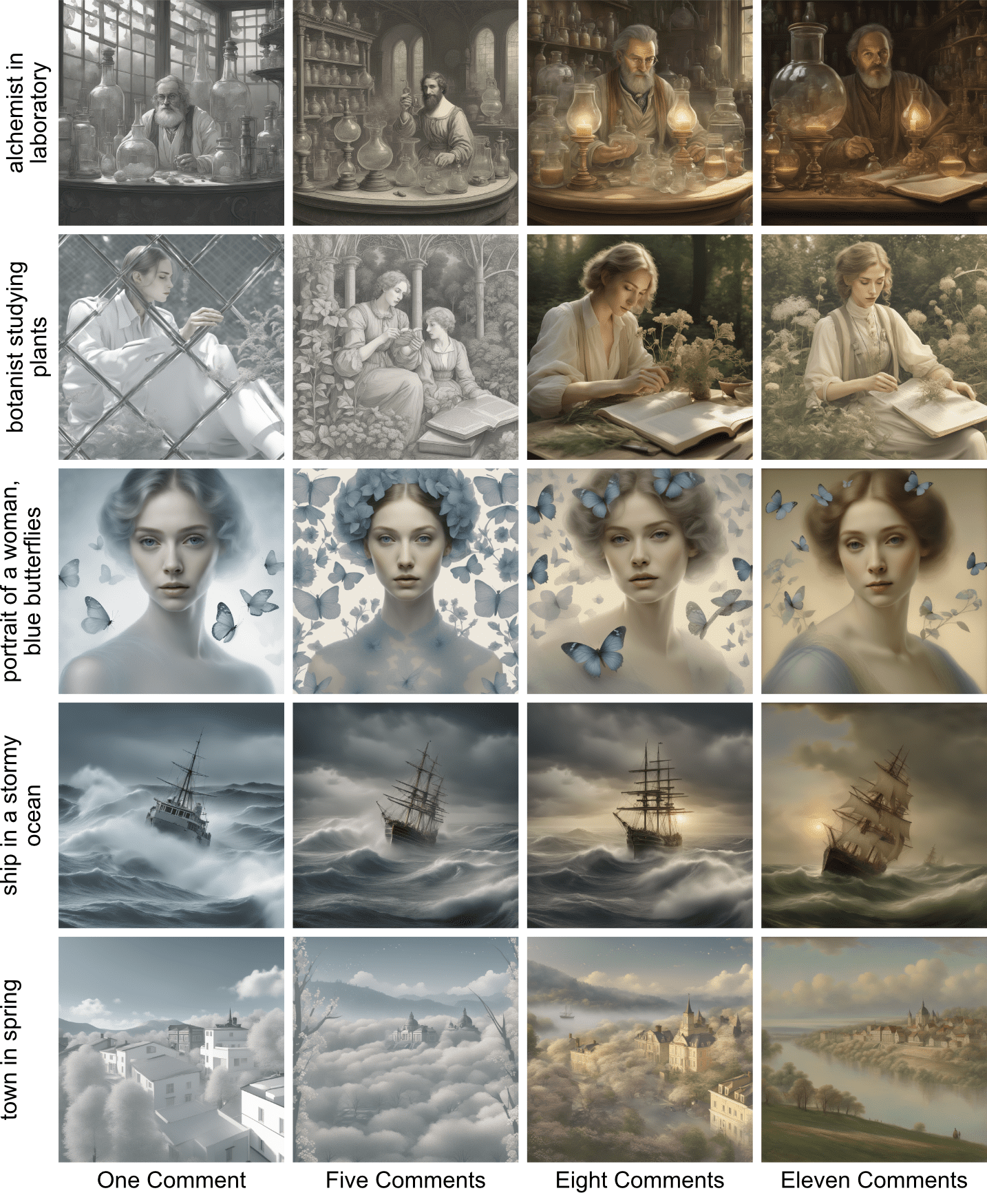}
  \caption{
  Each column is generated from visual preferences extracted from the same set of comments. In the leftmost column, we selected only one random comment, while in the rightmost column, we generated the visual preference from all the user's provided comments. The prompts are displayed on the left. It is evident that as the number of comments increases, the results become more aligned with the user's artistic preferences. While the first column only considered attributes such as distaste for "Abstract", "Contemporary", and "Vibrant Colors" along with a preference for "Delicate" and "Detailed" images, the second column included elements regarding color palette and historical vibe. These visual attributes expanded to the complete set presented in \cref{sec:numcom}.
  }
  \label{fig:numcomments}
\end{figure}
\label{sec:numcom}
After selecting which images to comment on, individuals are encouraged to discuss their likes or dislikes for each selected image, expressing their perception of it and the emotions it evokes. Importantly, individuals are not bound to any specific writing structure when commenting. They can write as much or as little as they prefer, as we utilize a language model to extract structured preferences from their free-form comments. Ideally, these comments are descriptive, covering most of the attributes present in the images that are important to the individual. We demonstrate that in this case, comments on as few as eight images are enough to obtain personalized results. 

Consider a user who has the following visual preference:

\textbf{Likes}: \textit{Transparent Texture, Glassy, Delicate, Detailed, Story-telling, Intricate Lines, Smooth Texture, Muted Earthy Colors, Historical, Realistic Lighting, Classicism, Busy Composition, Neo-classicism, Balanced, Open Space, Matte, Depth, Matching Colors, Renaissance, Blending Colors, Subtle Lighting, Ethereal, Atmospheric Perspective, Illusion of Depth, Oil Painting, Creativity, Soft Texture, Romanticism, Desaturated, Majestic, Watercolor}

\textbf{Dislikes}: \textit{Abstract, Contemporary, Vibrant Colors, Incoherent Shapes, Chaotic Brushstrokes, Simplistic, Blood and Gore, Bright and Blunt Colors, Single-focused, Realism, Flat, Polished, Modern Architecture, Extreme Colors, Monochromatic, Cold Colors, Picasso, Geometric Patterns, Somber, Dark, Dominating Red, Dry Brushstrokes, Monochromatic, Minimalism, Artificial Colors, Color Splash, Photorealistic, Cartoon, Childish}

In \cref{fig:numcomments} we randomly select a subset of this user's comments and extract their visual preferences from them. In the leftmost column, where only one random comment was selected, the results do not exactly align with the user's preferences in images. As the number of comments increases, they converge towards their general preference for classic art and related visual attributes, and with eight comments, the results are almost perfectly aligned with their taste, though the user still preferred the last column consistently. It is notable that this experiment is also dependent on the random subset chosen. If the user's selected comment in the first column indicated their extreme preference toward "Classicism", "Renaissance", or "Muted Earthy Colors", etc., the results would have aligned more with their preference. To avoid cases where commenting on an image fails to give us a lot of information about the user, we ask users to provide an average number of eight to ten comments on images that provoke strong reactions from them.

\subsection{Prompt Personalization}\label{sec:promptpers}

This approach involves modifying prompts using language models based on an individual's visual preference signal. We instruct the language model to adjust a given prompt based on the provided visual preference text. The objective is to obtain a tailored prompt that, when given to a generative model, can produce a personalized image. This approach serves as a baseline to explore another method of utilizing the $VP$ signals aside from ViPer's approach. The input prompt to the language model is as follows:\\

\begin{lstlisting}
I will provide you with a base prompt for image generation and a set of visual attributes preferred and a set of visual attributes disliked by a person. Your task is to personalize the base prompt by incorporating the person's preferred and disliked visual attributes such that it would consider the preferred attributes and avoid the disliked visual attributes. Your output should only be the personalized prompt, without any additional text. 
The personalized prompt should:
1. Reflect the person's preference in images, without directly copying the attributes.
2. Follow the standard prompt format and be concise, using keywords and short key-sentences.
3. Maintain the same core content as the base prompt.
Here's the base prompt: {Prompt}
Here's the person's preferred attributes in images: {VP+}
Here's the person's disliked attributes in images: {VP-}
\end{lstlisting}

This method has some shortcomings due to being dependent on a language model. This may lead to lengthy prompts, disregarding parts of visual preferences, and adopting a conversational tone, even against explicit instructions. Additionally, the personalized prompts may fail to include key concepts from the original prompts, e.g., the prompt \textit{"Paper clouds, brown origami birds"} was transformed into \textit{"German Expressionism-inspired clouds float gracefully alongside birds in a sky painted with earthy tones and azure hues. The scene is illuminated by the golden hour, casting a surreal glow over the vibrant landscape. The birds soar amidst dynamic, fragmented clouds, creating a pensive yet lively atmosphere"} when using ChatGPT. This resulted in the loss of the instructed materials in the prompt, such as paper and the concept of using origami, as well as the color brown, for the sake of incorporating visual attributes the individual prefers.

\begin{figure}[H]
  \centering
  \includegraphics[width=\linewidth]{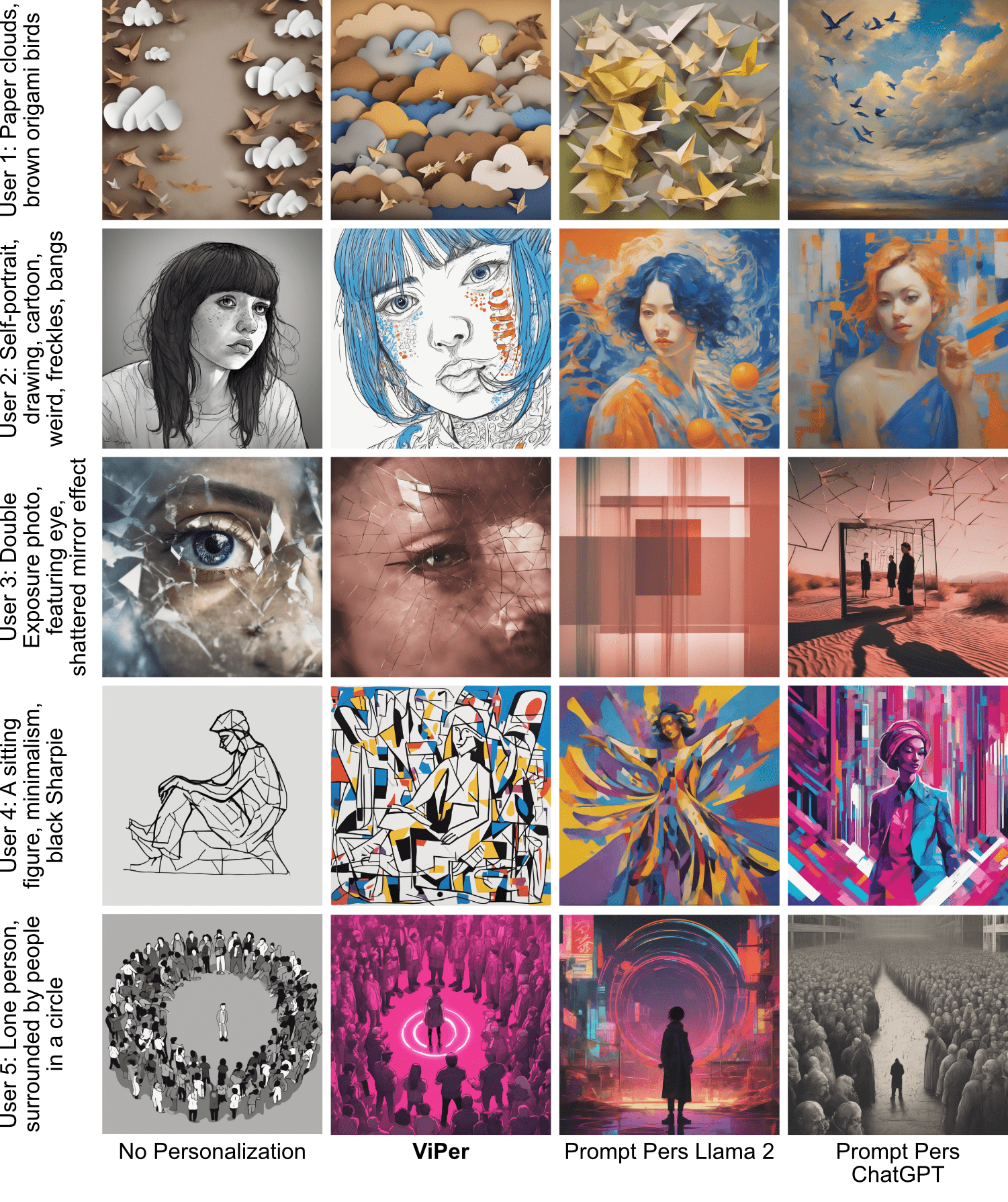}
  \caption{
Each row represents a different user and prompt, highlighting the shortcomings of Prompt Personalization. The input prompts are displayed on the left. The leftmost column presents generated responses without any personalization, while the subsequent three columns are personalized using either ViPer or Prompt Personalization with Llama 2 or ChatGPT. While losing the input prompt due to a lengthy personalized prompt is more common, the last row illustrates how ChatGPT completely ignored the user's visual preference (In this case, the personalized prompt became\textit{ Solitary figure stands amidst a gathering crowd}). In contrast, ViPer personalized the results while still adhering to the instructions provided in the input prompt. It's also notable that sometimes, the language model gets fixated on a specific visual attribute, leading to extreme results such as Llama 2's third row, which totally lost the concept of the input prompt by emphasizing minimalism, even though it was one of the many visual attributes present in the user's preference.
  }
  \label{fig:ppvsg}
\end{figure}

We show some of these samples in \cref{fig:ppvsg}. Each row represents a user's preference. We can see that ViPer considers the input prompts' consistently. However, depending on the language model in use, Llama 2-70b or ChatGPT, Prompt Personalization either misses the input prompt or the visual preference. Note that one cannot simply rely on appending the visual preference text to the input prompt, as Stable Diffusion would ignore most parts of the visual preference due to the length of the prompt.

\subsection{Consistency with Input Prompts}\label{sec:consistency}
\begin{figure}[h!]
  \centering
  \includegraphics[width=\linewidth]{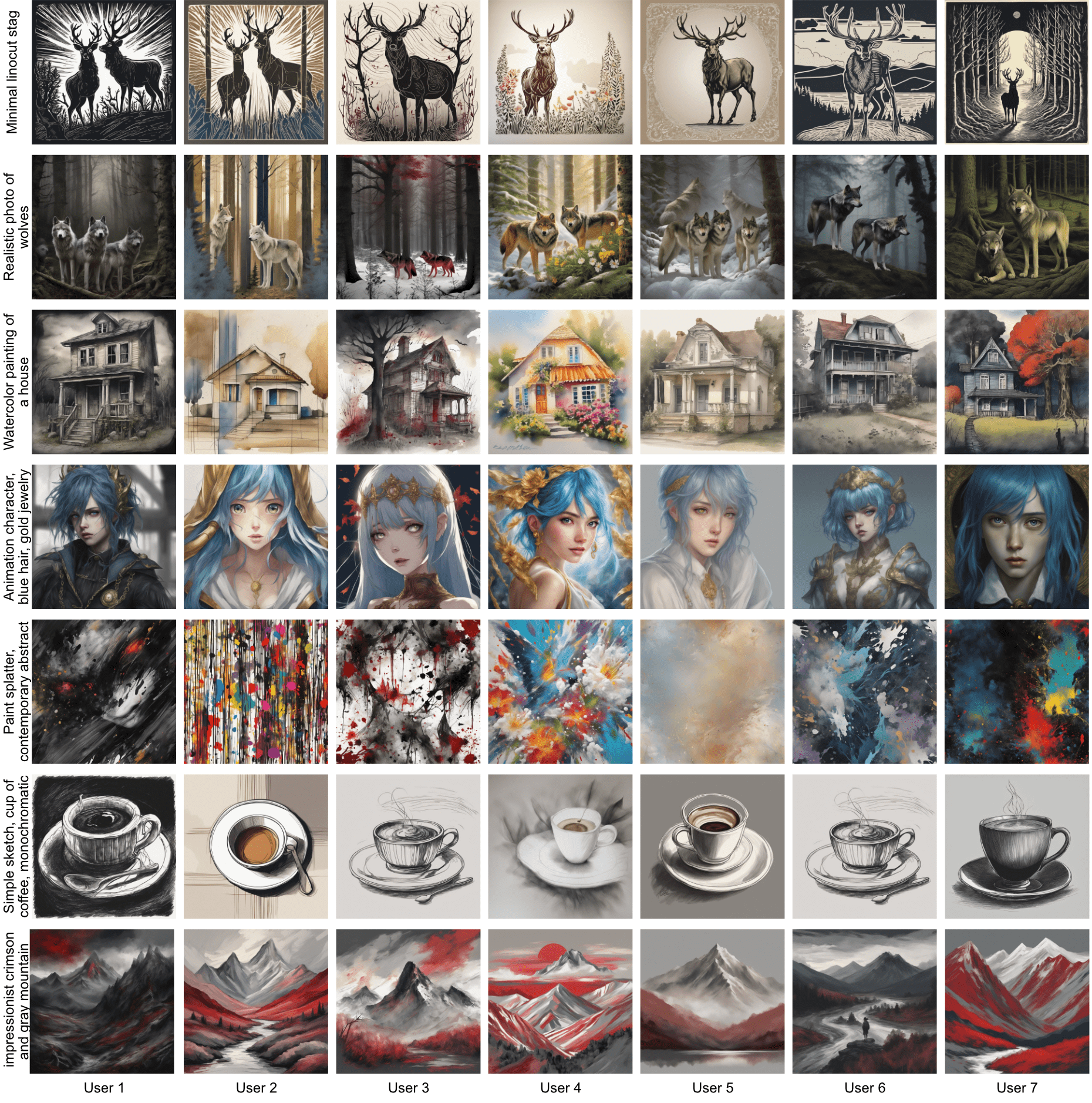}
  \caption{
  Each column is generated from a user's preference using ViPer, and the prompts are shown on the left. While the results are still somewhat personalized, they follow the visual attributes defined in the prompts.
  }
  \label{fig:stableprompt}
\end{figure}
\begin{figure}[t]
  \centering
  \includegraphics[width=1.\linewidth]{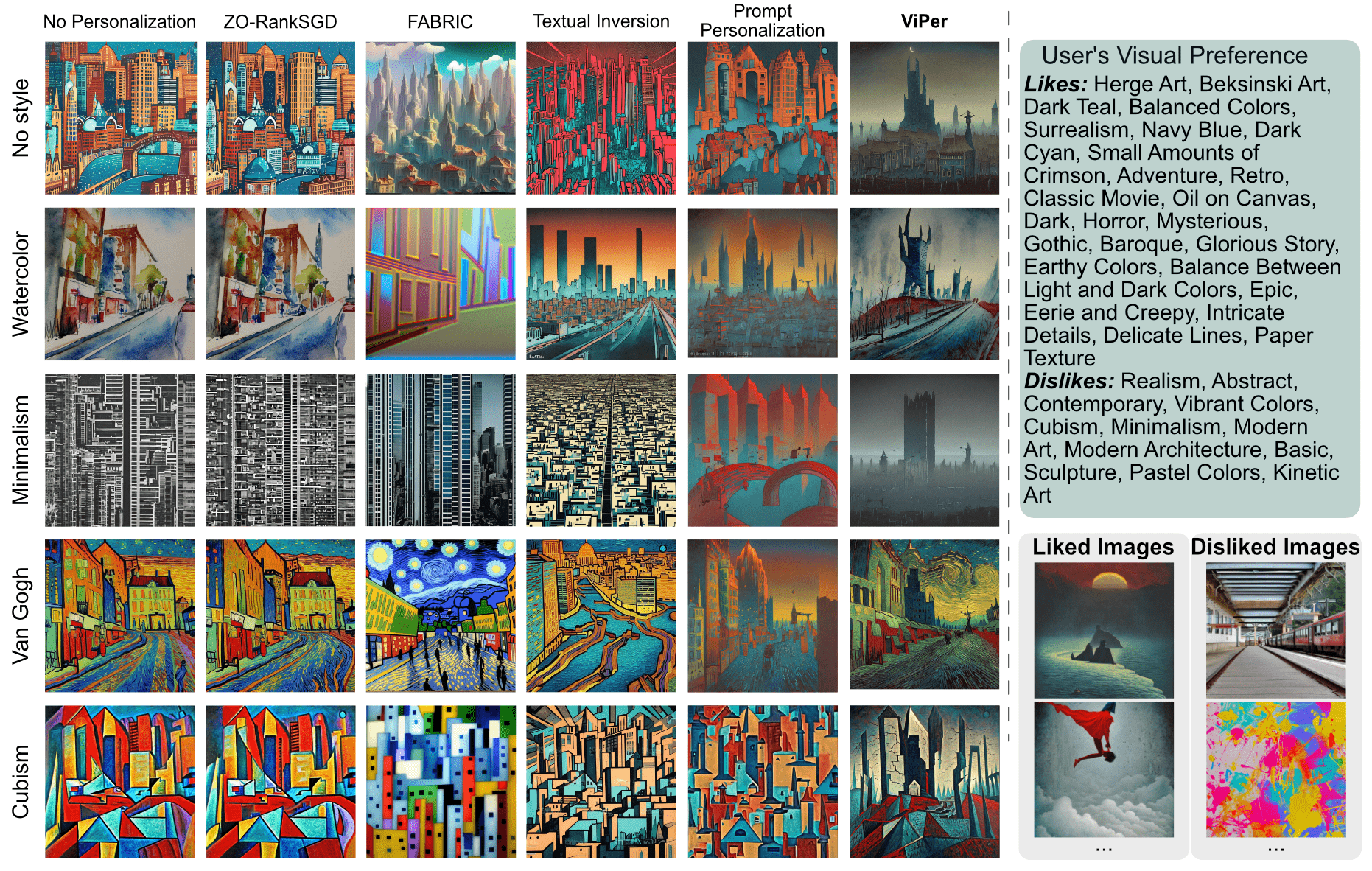}
  \caption{
  For the "Cityscape" prompt, we're comparing ViPer with baselines with specified styles in prompts. Each row follows the style on the left. The leftmost column has no personalization, while the next five columns show different methods' generations. ZO-RankSGD's results resemble those without personalization. FABRIC changes images, but the user doesn't find them personalized and they sometimes lose the original style (see row 2 in column 3). Textual inversion's results are too vibrant for the user's taste and lose the specified style (see row 4 in column 4). Prompt Personalization misses the "Van Gogh" style but aligns well with the user's preference. ViPer follows both the prompt's style and the user's visual preference. Note how the user dislikes "Cubism," yet it appeared in the result (see row 5 in column 5) when specified in the prompt. The user's visual preference and a small set of images they like and dislike are presented at right.
  }
  \label{fig:baselines}
\end{figure}

In our tests, input prompts dictate results, with visual preference embeddings limited by $\beta$ to prevent overshadowing prompt embeddings. In \cref{fig:stableprompt}, we present the results of seven users for seven prompts that explicitly define the art style, medium, or color palette. We can see that even though the results are still personalized and there is a clear preference pattern in each column, they consistently adhere to the input prompt's meaning. In \cref{fig:baselines} we compare the baseline results for a single prompt when the style is specified. Note how ZO-RankSGD’s \cite{tang2024zerothorder} generations are similar to the generations without any personalization. While FABRIC \cite{vonrütte2023fabric} changes the images, the results are not only not personalized but also sometimes lose the prompt's original style (see row 2 in column 3). Textual Inversion's \cite{gal2022image} results are too vibrant for the user's taste and lose the prompt's specified style as well (see row 3 and 4 in column 4). Prompt Personalization misses the "Van Gogh" style but is mostly aligned with the user's preference. However, ViPer follows both the input prompt's style and the user's visual preference. It's important to note that since the effect of input prompt is more than visual preferences, due to $\beta$'s limitation, individuals can generate images with attributes they dislike by mentioning them in the prompt (see row 5 in column 5).

\subsection{Effect of Visual Attributes}\label{sec:visattreffec}
\begin{figure}[h!]
  \centering
  \includegraphics[width=1.\linewidth]{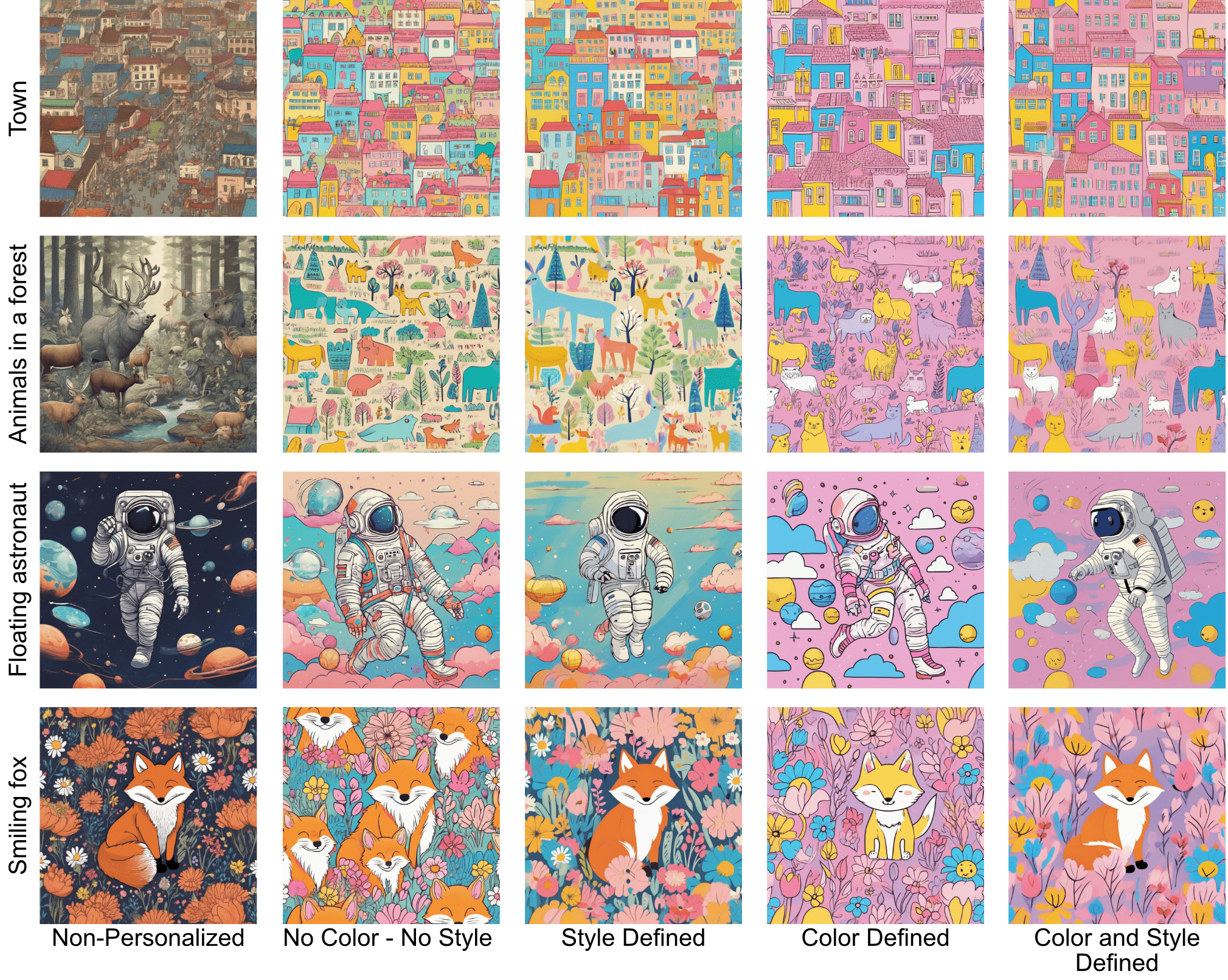}
  \caption{For an individual with preference for \textit{Bold Outlines, Sharpie, High-contrast, Simplified, Color-pencil, Kawaii, Alex Katz Style, Iwona Lifsches Style, Cartoon, Comic, Naive Art, Bright Colors, Playful, Pastel Palette, Paper Texture, Sketch Lines, Gouache, Opaque Watercolor, Nostalgic, Pink, Yellow, Blue, Lilac}, in each column we remove either the color palette attributes, art style attributes, or both, to show the effect of the other visual attributes in more detail. See page \pageref{visat} for the grouping and definition of visual attributes. There are 20 categories of attributes in general. Excluding color palette and art style, these categories include composition, mood, etc. The first column shows the non-personalized generations. Note how in the other columns, the overall pattern of the images remains consistent. This consistency arises because many visual attributes already establish the foundation of the preferred art styles and colors. Clearly, these effects vary for each unique visual preference, depending on the preferred attributes.}
  \label{fig:attrs}
\end{figure}

While visual attributes such as color palette and art style are immediately apparent to the eye upon first glance at images, it's important to note that other visual attributes also exert a significant effect on the results. To analyze this, we create 100 agents within three categories: those without preferred colors, those without preferred art styles, and those without both colors and styles. We then conduct an identifiability test on these agents, using our proxy metric to observe how distinguishable the results are. In this test, we initially generate images based on each agent's preferences. We then use these generated images as examples of images the agent likes, while selecting random images from other agents' preferences to represent images the agent doesn't like. Our expectation is that our proxy will assign higher scores to the images generated specifically for each agent and lower scores to the random images. We show the results of the test in \cref{tab:visattr}. It's evident that defining colors or art styles enhances our proxy's ability to better distinguish between liked and disliked images; however, other visual attributes are also undoubtedly important in this process. We show an individual's results in \cref{fig:attrs}. Notice how, even without defining specific colors or art styles, the overall pattern of the images remains consistent and all of the generations are personalized.

\begin{table}[h!]
\caption{To analyze the effect of different visual attributes, we study 100 agents divided into three groups: without preferred colors, without preferred art styles, and without both colors and styles. Results show that specifying color or art style helps our metric identify liked and disliked images better, but other visual attributes are also significant.} 
  \label{tab:visattr}
  \centering
\begin{tabular}{@{}c|c|c|c|c|c@{}}
\toprule

         & No Color - No Style & Style Defined & Color Defined & Both defined & Random\\ \midrule
Accuracy & 71.6\%              & 78.8\%        & 83.1\%  & 91.4\% & 50.0\% \\\bottomrule  
\end{tabular}
\end{table}

\subsection{Diversity Concerns}\label{sec:diversity}
\begin{figure}[H]
  \centering
  \includegraphics[width=\linewidth]{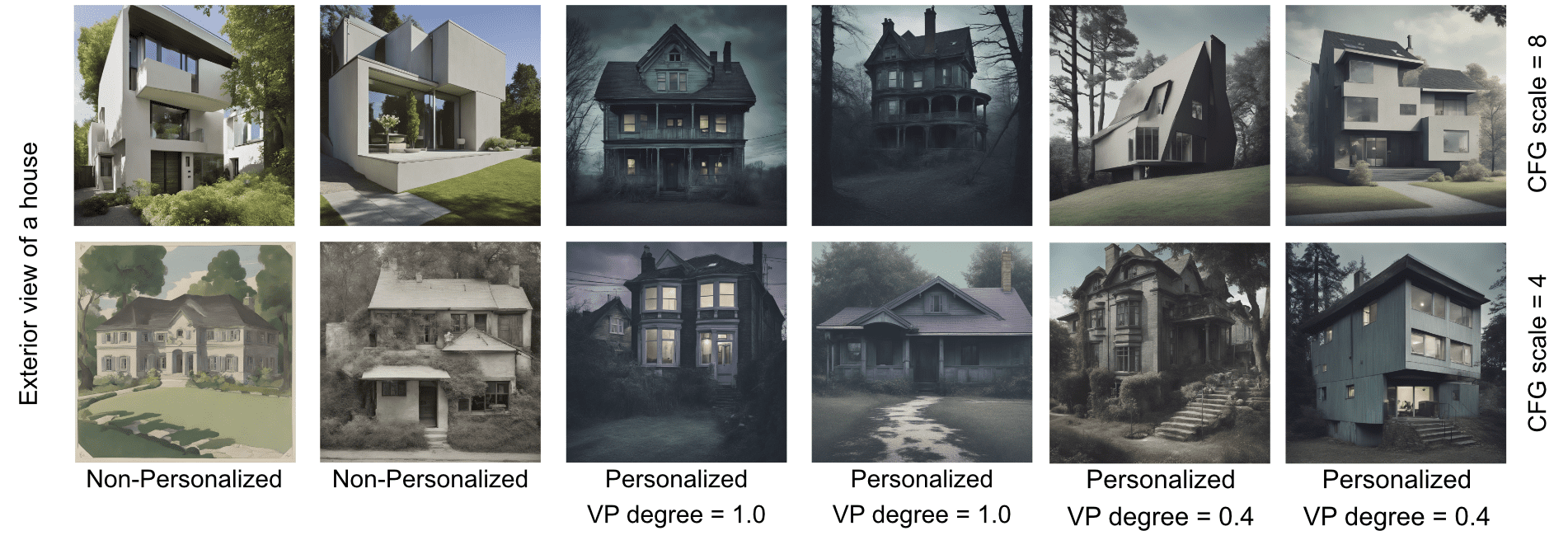}
  \caption{We generate multiple images from the same prompt with different $CFG$ scales and personalization degrees. Note how the outputs become more similar as the $CFG$ scale and personalization degree increase. One can increase the level of diversity by lowering these two hyperparameters.
  }
  \label{fig:diversityconcerns}
\end{figure}

One potential concern is that the proposed system might cause the diversity of outputs to collapse for a particular prompt. To address this, we compared the diversity of outputs from non-personalized and personalized models on the same prompt. ViPer allows users to adjust diversity through the personalization degree, similar to the classifier-free guidance ($CFG$) scale (see Fig. 5 of the main paper). Non-personalized Stable Diffusion tends to produce similar outputs with high $CFG$ scales, indicating a trade-off between prompt-image alignment and diversity. Similarly, ViPer presents a trade-off between diversity and personalization, modifiable by the personalization degree. By adjusting the personalization degree, users can control the balance between the specificity of the personalization and the diversity of the generated outputs. In \cref{fig:diversityconcerns} we show some images generated from different $CFG$ scales and personalization degrees. We can see that as we decrease these hyperparameters, the outputs become less similar.

\subsection{Effect of Number of Images on Proxy}
\begin{figure}[h!]
  \centering
  \includegraphics[width=0.8\linewidth]{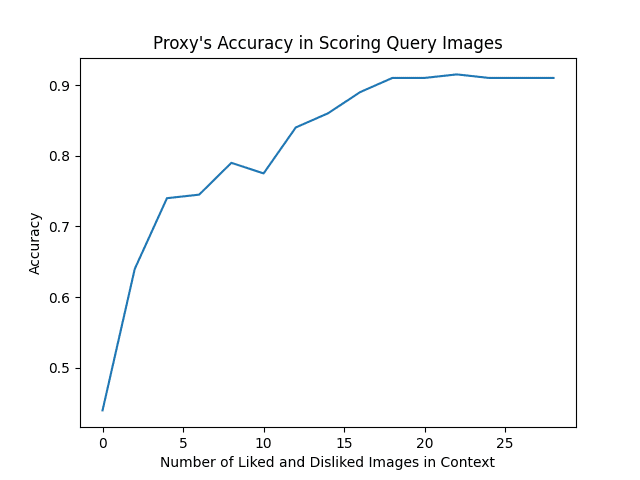}
  \caption{We generate 100 agents and 10 images based on each of their visual preferences. Then, for each agent, we conduct a test with our proxy $M$ to determine if it can assign a higher score to an agent's personalized images compared to random images, while varying the number of input images (both personalized and non-personalized). We observe that as the number of images in the context increases, the accuracy of the $M$ also increases, stabilizing at 91.1\% after approximately 18 input images.
  }
  \label{fig:proxypairs}
\end{figure}
To understand the impact of the number of images (both personalized and non-personalized for each agent) in the input prompt for the proxy, we evaluate the proxy's ability to distinguish between an agent's personalized and non-personalized query images. This evaluation involves providing the proxy with a varying number of personalized and non-personalized images (ranging from 0 to 28) as context in the input prompt.

For each agent, we measure accuracy based on whether the proxy assigns a higher score to the personalized query compared to a non-personalized image. In \cref{fig:proxypairs}, we depict the relationship between accuracy and the number of input images in the context by testing 100 agents. We can see that with fewer input images, the metric's predictions are closer to random. However, as we increase the number of images in the input prompt, the proxy's accuracy improves for both test sets, stabilizing at approximately 91.1\% accuracy after around 18 images in the context. This indicates that with more data (i.e., more input images), the proxy gains a better understanding of the agents' visual preferences, resulting in more reliable outcomes.

\subsection{Fine-Tuning Stable Diffusion with Individual Preferences} 
\begin{figure}[h!]
  \centering
  \includegraphics[width=\linewidth]{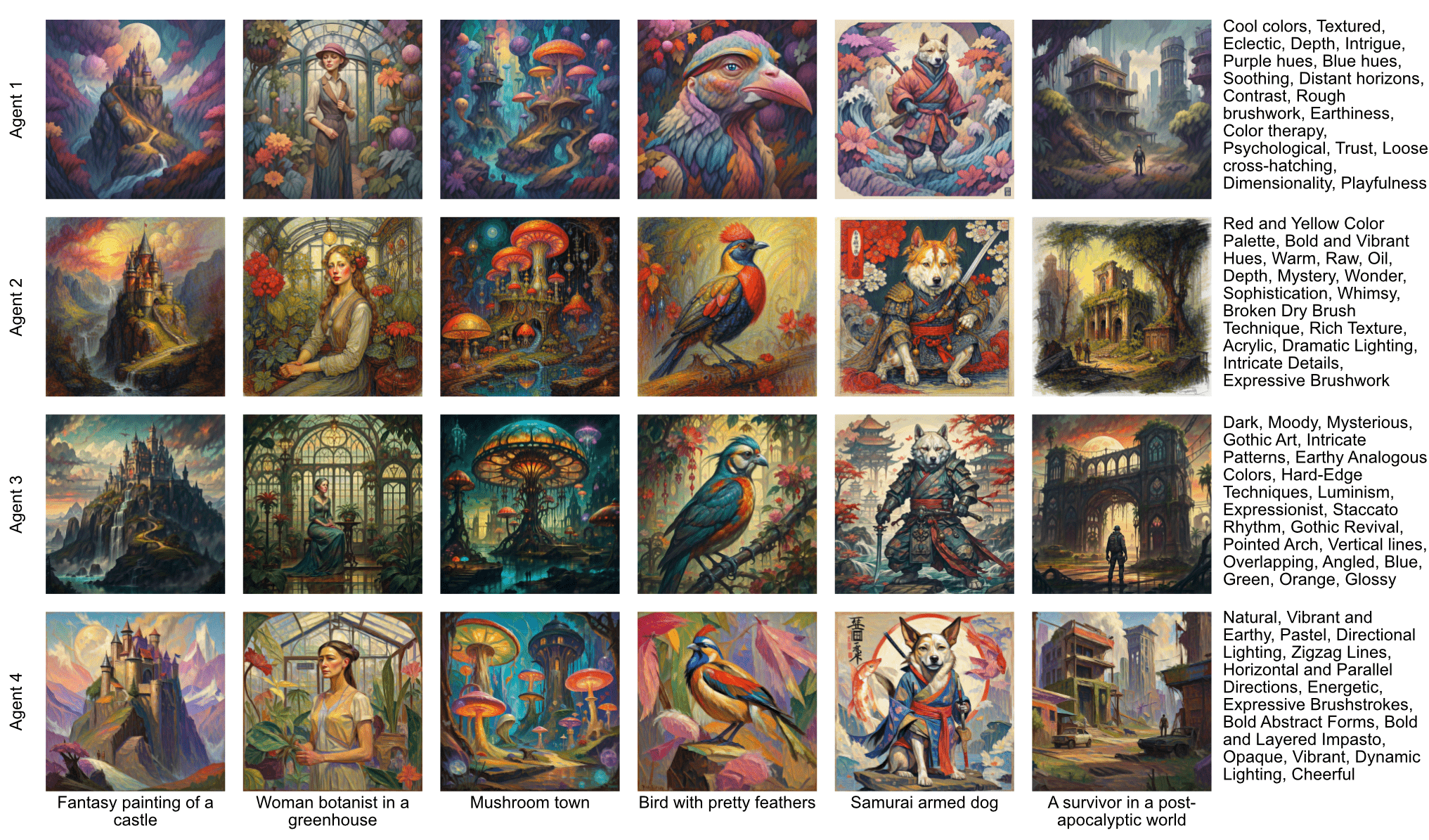}
  \caption{In this experiment, we fine-tune Stable Diffusion on each agent's visual preferences using the CLIP similarity score for the generated images at each step and the visual preference texts. Note how, for each agent, the results align with the visual preference texts shown at right.
  }
  \label{fig:finetune}
\end{figure}

To align Stable Diffusion to users' visual preferences, we conducted experiments involving fine-tuning the model based on each individual's preferences. We used the CLIP similarity score in this experiment, comparing Stable Diffusion-generated outputs for each user against their specified visual preference text to serve as a reward model \cite{xu2023imagereward, clark2023directly}. In \cref{fig:finetune}, we showcase the outcomes of this approach for four distinct agents, with each row depicting personalized images corresponding to an agent's visual preference (displayed on the right), and each column containing images generated from the same prompt.

While the results of this approach were promising, it exhibited several shortcomings compared to ViPer. Notably, this method is quite expensive due to the necessity of tuning Stable Diffusion for each user.  Moreover, it's sensitive to over-optimization (reward-hacking)\cite{clark2023directly}, resulting in loss of diversity and collapsing as training steps increase. In severe instances, the tuned model disregards input prompts entirely, only to output the same high-reward images. In milder cases, it generates the same facial features, textures, or extreme patterns, resulting in lower-quality images. Additionally, this method may lead to certain visual attributes being overlooked in favor of emphasizing those that yield higher scores from the reward model.


\section{Broader Impact}\label{sec:broadimpact}
\begin{figure}[H]
  \centering
  \includegraphics[width=\linewidth]{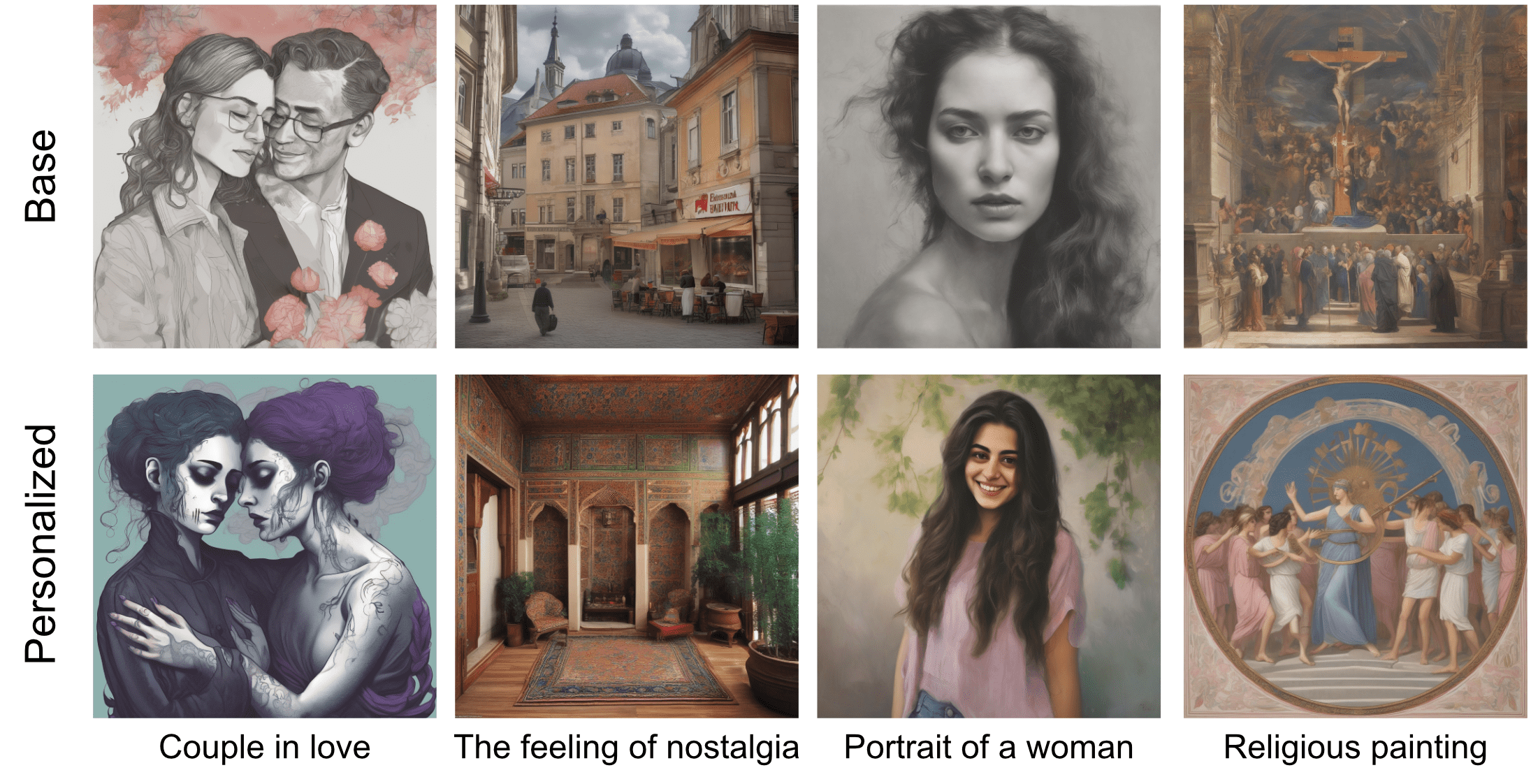}
  \caption{ We present the non-personalized (base) results of standard Stable Diffusion in the top row. The bottom row displays personalized samples generating the same concepts for different users. While standard Stable Diffusion often discards diversity due to its training data, our method provides individuals with the opportunity to express themselves and their experiences, which are reflected in their visual preferences.
  }
  \label{fig:diversity}
\end{figure}

Visual preferences vary widely based on personal experiences and backgrounds. However, current generative models often fail to capture this diversity due to limited training data, resulting in generic outputs. For example, prompts like \textit{nostalgia} may lean towards Western aesthetics, \textit{love} often depicts heterosexual monogamous couples, \textit{religion} incorporates Christian symbols, and \textit{human} may exclude minority races. Our method addresses these limitations by allowing users to express their preferences thoroughly and in various concepts. In our user tests, some participants favored Queer art and depicted gender nonconforming roles, highlighting the need for more inclusive representation in generative models. Our approach easily incorporates these preferences without requiring additional training data or constant prompt tuning from individuals. Some of these samples are shown for different individuals in \cref{fig:diversity}.

\section{Human Subjects Data}\label{sec:humandata}
\begin{figure}[h!]
  \centering
  \includegraphics[width=\linewidth]{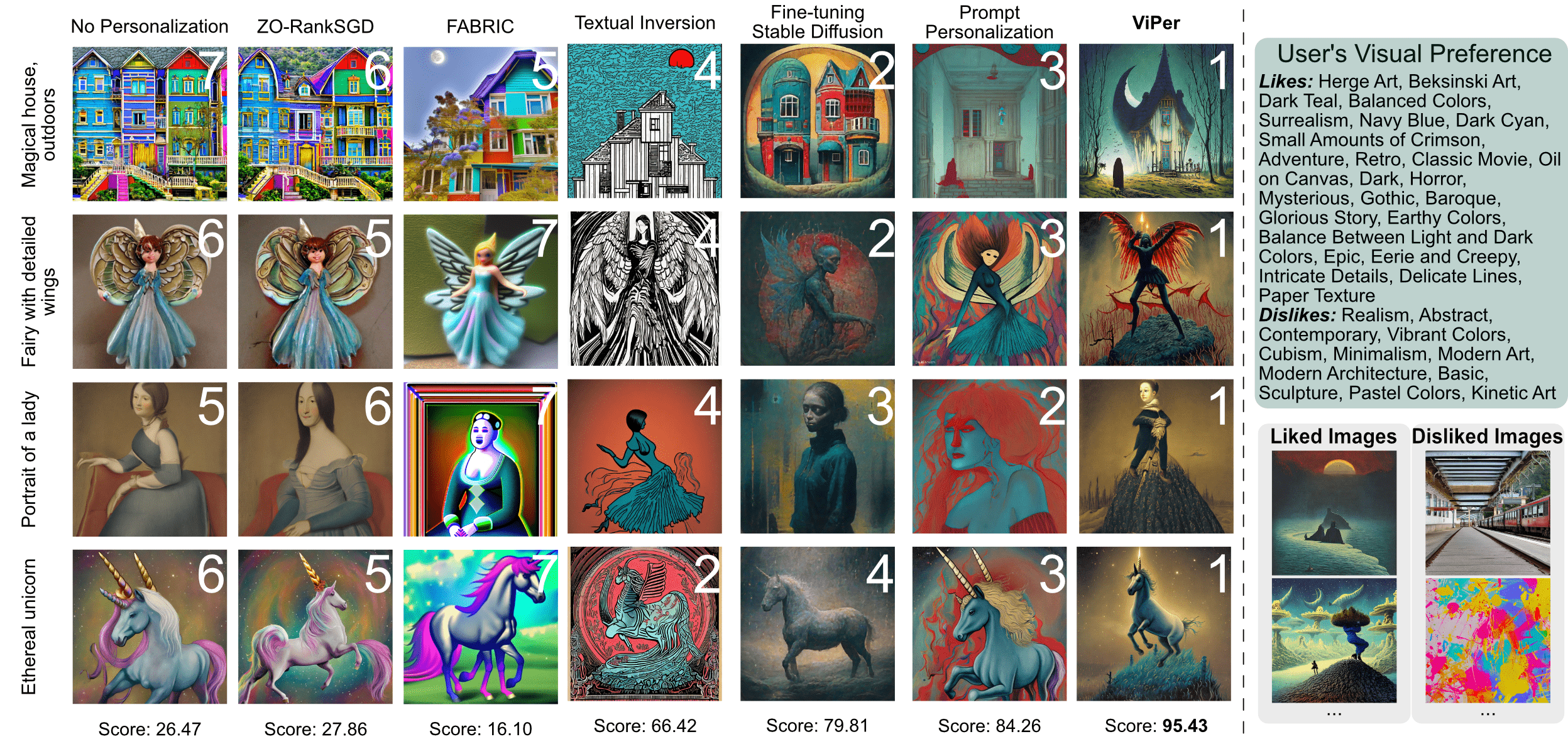}
  \caption{ We present the user's ranking in Fig. 6 of the main paper. For each prompt, the user ranked the images based on their preference. They consistently preferred ViPer over other baselines. Note how the user's rankings for different baselines align with the average scores of metric $M$.
  }
  \label{fig:ranking}
\end{figure}
We only use human subjects to evaluate the accuracy of our method and its alignment with actual human preferences. This data contains no identifiable information, and whenever an annotation or a part of a user's data was solicited, their consent was requested. In \cref{tb:VPS}, we show the visual preferences of different individuals shown in Fig. 4 of the main paper. The user's ranking in Fig. 6 of the main paper is presented in \cref{fig:ranking}.

\begin{table}[htbp]
\centering
\caption{Visual preferences of different users shown in Fig. 4 of the main paper.}
\adjustbox{max width=\textwidth}{%
\begin{tabular}{p{2.5cm}|p{5cm}|p{5cm}} \toprule
\textbf{User} & \textbf{Likes} & \textbf{Dislikes} \\ \midrule
1 & \begin{tabular}[t]{@{}p{5cm}@{}}Dark, Edgy, Detailed, Busy Composition, Eye Symbolism,  Eerie, Creepy, Soft Brushstrokes, Raw, Unsettling, Surrealism, Chiaroscuro Lighting, Monochromatic, Ominous, Symbolic,  Horror, Grey  Colors, Illusory, Distorted Perspective, Dramatic, Theatrical, Fluid Lines, Desaturated, Classic\end{tabular} & \begin{tabular}[t]{@{}p{5cm}@{}}Contemporary Abstraction, Cheerful Colors, Colorful, Cartoonish, Anime, Manga, Basic, Hyper-realistic, Minimalism, Bright, Flat, Cubism,  Repetitive Patterns, Pops of Color, Vibrant Hues, Art Deco, De Stijl, Pixel Art\end{tabular} \\ \midrule
2 & \begin{tabular}[t]{@{}p{5cm}@{}}Realism, Representational, Vibrant, Lighter Colors, Bright, Light and Vivid Shades, Detailed, Cheerful, Positive, Uplifting, Soft, Natural Elements, Organic, Atmospheric Perspective, Hopeful,  Harmony, Figure-Ground Relationship, Open Space Composition, Dynamic, Colored Pencil\end{tabular} & \begin{tabular}[t]{@{}p{5cm}@{}}Melancholic, Gore, Muted Colors, Dark Themes, Monochromatic, Graffiti, Bad Painting, Abstract, Ambiguous, Artificial Hues, Grayscale, Fog Lighting, Collage, Linocut, Woodcut,  Cyanotype Photography, Black and White, Harsh Lighting\end{tabular} \\ \midrule
3 & \begin{tabular}[t]{@{}p{5cm}@{}}Psychological and Emotional Depth, Queer Art, Sketch Lines, Dark Themes, Magical, Gloomy, Melancholy, Various Color Palettes, Eerie Vibes, Melancholic, Hand-drawn Textures, Dramatic Lighting, Storytelling, Gothic, Surrealism, Expressionism, Fantastical Elements, Somber Colors, Complex Composition, Innovative, Intricacy, Wonder, Fluid, Freehand Lines, Purples and Blues\end{tabular} & \begin{tabular}[t]{@{}p{5cm}@{}}Realism, Cheerful Colors, Vibrant, Minimalism, Contemporary Abstract, Yellow Hues,  Direct Lighting, Monochromatic, Black and white, Grayscale, Static, Lyrical Abstraction, Art Informel, Naive, Rough Texture, Strict, Straight and Thin Lines, Photography, 3D\end{tabular} \\ \midrule
4 & \begin{tabular}[t]{@{}p{5cm}@{}}Minimalist, Abstract, Simple Compositions, Room for Interpretation, Limited Color Palettes, Flat,  Subtlety, Open Spaces, Muted Colors, Desaturated Palettes, Monochromatic, Complementary Colors,  Aqueous Texture, Washed Out, Soothing, Subtle Texture, Contemporary Abstraction, Buttercream Hues, Harmonious, Compressed, Measured\end{tabular} & \begin{tabular}[t]{@{}p{5cm}@{}}Detailed, Busy Composition, Vibrant Colors, Realism, Baroque, Neo-Impressionism, Vivid, Complex, Intricate details, Mechanical Lines, High Contrast, Chaotic, Maximalist  Juxtaposition\end{tabular} \\ \bottomrule
\end{tabular}}
\label{tb:VPS}
\end{table}

\FloatBarrier
\afterpage{
\noindent\textbf{Visual Attributes}\label{visat}\\

\scriptsize

\noindent\textbf{Art styles:}
\noindent\begin{center}\begin{minipage}{0.95\linewidth}
\texttt{1. Aboriginal Art 2. Abstract Expressionism 3. Academic Art 4. Ancient Greek Art 5. Anime, Manga 6. Art Deco 7. Art Informel 8. Art Nouveau 9. Art of Ancient Egypt 10. Arte Povera 11. Assemblage 12. Bad Painting 13. Baroque Art 14. Bauhaus art 15. Byzantine Art 16. COBRA art 17. Ceramic Art 18. Chinese Art 19. Color Field Painting 20. Conceptual Art 21. Cubism 22. Dada 23. De Stijl 24. Digital Art 25. Fauvism 26. Feminist Art 27. Futurism 28. Queer Art 29. German Expressionism 30. Gothic Art 31. Graffiti Art 32. Impressionism 33. Indian Art 34. Installation Art 35. Islamic Art 36. Japanese Art 37. Kinetic Art 38. Land Art 39. Lyrical Abstraction 40. Magic Realism 41. Mannerism 42. Medieval Art 43. Mesoamerican Art 44. Metaphysical Painting 45. Minimalism 46. Naive Art 47. Native American Art 48. Neo-Impressionism 49. Neoclassicism 50. New Media Art 51. Oceanic Art 52. Op Art 53. Performance Art 54. Photorealism 55. Pixel Art 56. Pointillism 57. Pop Art 58. Postmodern Art 59. Realism 60. Regionalism Art 61. Renaissance Art 62. Rococo Art 63. Romanticism 64. Situationist Art 65. Social Realism 66. Street Art 67. Stuckism 68. Surrealism 69. Symbolism 70. Traditional African Art 71. Ukiyo-e 72. Video Art 73. Contemporary Abstraction}
\end{minipage}\end{center}

\noindent\textbf{Color palette:}
\noindent\begin{center}\begin{minipage}{0.95\linewidth}
\vspace{-1em}
\texttt{\begin{itemize}
\item Vibrant: 1. Electric Blue 2. Radiant Red 3. Citrus Orange 4. Lively Yellow 5. Vivid Purple 6. Hot Pink
\item  Muted Tones: 1. Dusty Rose 2. Soft Sage 3. Faded Denim 4. Subtle Gray 5. Antique Lavender
\item Pastel Hues: 1. Powder Blue 2. Mint Green 3. Blush Pink 4. Lilac 5. Peach 6. Buttercream
\item Earthy Colors: 1. Olive Green 2. Terracotta 3. Warm Brown 4. Deep Forest Green 5. Rustic Red 6. Clay Gray
\item Neon Shades: 1. Electric Lime 2. Pink 3. Fluorescent Orange 4. Laser Blue 5. Atomic Green 6. Hot Magneta
\item Monochromatic: 1. Charcoal Gray 2. Slate Blue 3. Steel Gray 4. Midnight Black 5. Silver White 6. Graphite
\item Oceanic Tones: 1. Deep Sea Blue 2. Turquoise 3. Teal 4. Aquamarine 5. Navy 6. Ocean Green
\item Autumnal: 1. Burnt Orange 2. Harvest Gold 3. Olive Brown 4. Pumpkin Spice 5. Cranberry Red 6. Autumn Leaf
\item Desert Sunset: 1. Sandstone 2. Coral Pink 3. Dusty Orange 4. Warm Terracotta 5. Desert Rose 6. Goldenrod
\item Rain-forest: 1. Jungle Green 2. Fern Green 3. Mossy Brown 4. Orchid Purple 5. Tropical Teal 6. Earthy Amber
\item Urban Industrial: 1. Concrete Gray 2. Steel Blue 3. Urban Green 4. Iron Black 5. Yellow 6. Alloy Silver
\item Dark Colors: 1. Deep Indigo 2. Charcoal Black 3. Midnight Navy 4. Espresso Brown 5. Burgundy Wine 6. Forest Green
\item Black: 1. Jet 2. Obsidian 3. Onyx 4. Raven 5. Ebony 6. Charcoal
\end{itemize}}
\end{minipage}\end{center}

\noindent\textbf{Composition:}
\noindent\begin{center}\begin{minipage}{0.95\linewidth}
\texttt{1. Rule of Thirds 2. Golden Ratio 3. Symmetry/Asymmetry
4. Balance/Imbalance 5. Foreground/Background 6. Negative Space/Positive Space 7. Centralized/Decentralized 8. Grid-based 9. Contrasting Sections 10. Dynamic/Static 11. Figure-ground relationship 12. Shallow space/Deep space 13. Open space/Closed space 14. Real space/Invented space 15. Flatness/Illusion of Depth 16. Pictorial/Installation 17. Compressed/Exaggerated 18. Fragmented/Flattened 19. Rhythmic }
\end{minipage}\end{center}

\noindent\textbf{Skill:}
\noindent\begin{center}\begin{minipage}{0.95\linewidth}
\texttt{1. Masterful/Naive 2. Sophisticated 3. Expressive 4. Controlled 5. Gestural 6. Virtuosic 7. Restrained 8. Bold 9. Nuanced 10. Finessed 11. Heavy-handed 12. Subtle 13. Precise 14. Dynamic 15. Effortless 16. Inventive 17. Fluid 18. Intuitive 19. Meticulous 20. Robust 21. Graceful 22. Spontaneous 23. Deliberate 24. Energetic 25. Measured 26. Eloquent 27. Unconventional 28. Polished/Raw 29. Impulsive 30. Adaptive 31. Stylized 32. Rigorous 33. Efficient 34. Experimental 35. Graceful/Powerful 36. Unpredictable/Consistent 37. Free-flowing/Structured 38. Lyrical/Punchy 39. Innovative/Classic}
\end{minipage}\end{center}

\newpage

\noindent\textbf{Detail:}
\noindent\begin{center}\begin{minipage}{0.95\linewidth}
\texttt{1. Fine 2. Intricate 3. Selective 4. Soft 5. Textured 6. Tactile 7. Simplified 8. Focused 9. Blurred/Unfocused 10. Realistic 11. Abstracted 12. Expressive 13. Smooth 14. Complex/Minimalistic 15. Elaborate 16. Subtle 17. Defined 18. Rough 19. Ambiguous 20. Sharp 21. Ethereal 22. Muted 23. Vivid}
\end{minipage}\end{center}

\noindent\textbf{Hues:}
\noindent\begin{center}\begin{minipage}{0.95\linewidth}
\texttt{1. Red 2. Burgundy 3. Cerulean 4. Scarlet 5. Orange 6. Slate 7. Teal 8. Copper 9. Yellow 10. Green 11. Emerald 12. Gold 13. Turquoise 14. Azure 15. Indigo 16. Slate Gray 17. Blue 18. Magenta 19. Rose 20. Crimson }
\end{minipage}\end{center}

\noindent\textbf{Saturation:}
\noindent\begin{center}\begin{minipage}{0.95\linewidth}
\texttt{1. Highly saturated 2. Desaturated 3. Subdued 4. Intense 5. Grayscale 6. Pops of color}
\end{minipage}\end{center}

\noindent\textbf{Line:}
\noindent\begin{center}\begin{minipage}{0.95\linewidth}
\texttt{1. Straight lines 2. Curved lines 3. Zigzag lines 4. Diagonal lines 5. Horizontal lines 6. Vertical lines 7. Jagged lines 8. Freehand lines 9. Mechanical lines 10. Continuous lines 11. Thick lines/Thin lines 12. Fluid lines 13. Smooth lines 14. Wavy lines 15. Broken lines 16. Expressive lines 17. Spiral lines}
\end{minipage}\end{center}

\noindent\textbf{Shape:}
\noindent\begin{center}\begin{minipage}{0.95\linewidth}
\texttt{1. Square 2. Circle 3. Ellipses 4. Triangle 5. Rectangle 6. Oval 7. Hexagon 8. Organic 9. Inverted Triangles}
\end{minipage}\end{center}

\noindent\textbf{Value:}
\noindent\begin{center}\begin{minipage}{0.95\linewidth}
\texttt{1. Light 2. Dark 3. Mid-tone 4. High contrast/Low contrast }
\end{minipage}\end{center}

\noindent\textbf{Pattern:}
\noindent\begin{center}\begin{minipage}{0.95\linewidth}
\texttt{1. Stripes 2. Dots 3. Grid 4. Chevron 5. Herringbone 6. Floral 7. Geometric 8. Abstract 9. Motifs 10. Rhythm 11. Repetitive 12. Variation 13. Ordered 14. Random 15. Organic 16. Uniform 17. Progressing 18. Dense 19. Sparse 20. Chaotic 21. Intricate 22. Half-tone}
\end{minipage}\end{center}

\noindent\textbf{Texture:}
\noindent\begin{center}\begin{minipage}{0.95\linewidth}
\texttt{1. Smooth 2. Rough 3. Bumpy 4. Soft 5. Hard 6. Glossy/Matte 7. Impasto 8. Crisp 9. Fibrous 10. Granular 11. Porous 12. Tactile 13. Crackled 14. Slick 15. Peeling 16. Translucent/Transparent}
\end{minipage}\end{center}

\noindent\textbf{Artistic Medium:}
\noindent\begin{center}\begin{minipage}{0.95\linewidth}
\vspace{-1em}
\texttt{
\begin{itemize}
    \item Painting: 1. Oil 2. Acrylic 3. Watercolor 4. Ink 5. Encaustic 6. Gouache 7. Tempera 8. Spray Paint 9. Fresco 
    \item Drawing: 1. Charcoal 2. Pastel 3. Colored Pencil 4. Graphite 5. Chalk 6. Conte 7. Carbon Pencil 8. Silverpoint 
    \item Mixed Media: 1. Collage 2. Found Object 3. Paper Mache 4. Textile 5. Decoupage 6. Assemblage 7. Digital Collage 
    \item Printmaking: 1. Engraving 2. Etching 3. Linocut 4. Woodcut 5. Monotype 6. Silkscreen 7. Lithography 8. Drypoint 
    \item Digital Art: 1. Digital Painting 2. Pixel Art 3. 3D Modeling 4. 2D 5. AI Art 6. Virtual Reality 7. Concept Art 
    \item Photography: 1. Black and White 2. Film 3. Polaroid 4. Photomontage 5. Photogram 6. Cyanotype 
    \item Sculpture: 1. Stone Carving 2. Wood Carving 3. Metal 4. Ceramic 5. Glass 6. Found Object 7. Kinetic 8. Environmental
    \item Textile Arts: 1. Fiber 2. Quilting 3. Embroidery 4. Weaving 5. Knitting 6. Crocheting 7. Macrame 8. Batik 
    \item Ceramics: 1. Pottery 2. Porcelain 3. Stoneware 4. Raku 5.Earthenware 6. Terra Cotta 7. Slipcasting 8. Coiling
\end{itemize}
}
\end{minipage}\end{center}

\noindent\textbf{Mood:}
\noindent\begin{center}\begin{minipage}{0.95\linewidth}
\texttt{1. Somber 2. Joyful 3. Serene 4. Mysterious 5. Dramatic 
6. Theatrical 7. Tense 8. Emotional 9. Ambiguous 10. Calm 11. Unsettling 12. Whimsical 13. Nostalgic 14. Energetic 15. Hopeful 16. Reflective 17. Uplifting 18. Melancholic 19. Soothing 20. Adventurous 21. Enigmatic 22. Dreamy 23. Triumphant 24. Festive 25. Spooky 26. Mellow 27. Intense 28. Pensive 29. Ethereal 30. Funky 31. Regal 32. Hypnotic 33. Sassy 34. Lighthearted 35. Grandiose 36. Epic 37. Melodic 38. Quirky 39. Dynamic 40. Intriguing 41. Tranquil 42. Pensive 43. Sensual 44. Mystical 45. Urgent 46. Lively 47. Gentle 48. Eccentric 49. Euphoric 50. Brooding 51. Fierce}
\end{minipage}\end{center}

\noindent\textbf{Perspective:}
\noindent\begin{center}\begin{minipage}{0.95\linewidth}
\texttt{1. Linear 2. Atmospheric 3. Multiple 4. Distorted}
\end{minipage}\end{center}

\noindent\textbf{Depth:}
\noindent\begin{center}\begin{minipage}{0.95\linewidth}
\texttt{1. Shallow 2. Deep 3. Layered 4. Flat 5. Illusory 6. Spatial}
\end{minipage}\end{center}

\noindent\textbf{Movement:}
\noindent\begin{center}\begin{minipage}{0.95\linewidth}
\texttt{1. Static 2. Flowing 3. Rhythmic 4. Radial 5. Diagonal 6. Implied Motion 7. Gestural 8. Energy 9. Calmness 10. Direction 11. Animation}
\end{minipage}\end{center}

\noindent\textbf{Form:}
\noindent\begin{center}\begin{minipage}{0.95\linewidth}
\texttt{1. 2D 2. 3D 3. Flat 4. Dimensional/Abstracted 5. Solid/Transparent 6. Ambiguous 7. Fragmented 8. Geometric/Organic 9. Distorted 10. Simplified 11. Exaggerated}
\end{minipage}\end{center}

\noindent\textbf{Juxtaposition:}
\noindent\begin{center}\begin{minipage}{0.95\linewidth}
\texttt{1. Unexpected/Harmonious 2. Contrasting 3. Conceptual 4. Dynamic 5. Subtle/Bold 6. Organic 7. Minimalist/Maximalist}
\end{minipage}\end{center}

\noindent\textbf{Iconography:}
\noindent\begin{center}\begin{minipage}{0.95\linewidth}
\texttt{1. Symbolic 2. Allegorical 3. Religious 4. Mythological 5. Contemporary 6. Archetypal 7. Narrative 8. Spiritual 9. Pop culture references 10. Cultural 11. Emotional 12. Metaphorical 13. Archetypes 14. Icons 15. Motifs 16. Hidden Meanings}
\end{minipage}\end{center}

\noindent\textbf{Brushstrokes:}
\noindent\begin{center}\begin{minipage}{0.95\linewidth}
\texttt{1. Impasto 2. Drybrush 3. Wash 4. Blending 5. Cross-hatching 6. Scumbling 7. Sgraffito 8. Stippling 9. Wet-on-wet 10. Glazing 11. Palette knife 12. Feathering}
\end{minipage}\end{center}
}
\normalsize

\end{document}